# LNMR 2013

First International Workshop on
Learning and Nonmonotonic Reasoning

Post-Proceedings

September 15, 2013, Corunna, Spain
in association with LPNMR 2013

# Preface

Knowledge representation and reasoning (KR&R) and machine learning are two important fields in artificial intelligence (AI). (Nonmonotonic) logic programming (NMLP) and answer set programming (ASP) provide formal languages for representing and reasoning with commonsense knowledge and realize declarative problem solving in AI. On the other side, inductive logic programming (ILP) realizes inductive machine learning in logic programming, which provides a formal background to inductive learning and the techniques have been applied to the fields of relational learning and data mining. Generally speaking, NMLP and ASP realize nonmonotonic reasoning while lack the ability of (inductive) learning. By contrast, ILP realizes inductive machine learning while most techniques have been developed under the classical monotonic logic. With this background, some researchers attempt to combine techniques in the context of nonmonotonic inductive logic programming (NMILP). Such combination will introduce a learning mechanism to programs and would exploit new applications on the NMLP side, while on the ILP side it will extend the representation language and enable to use existing solvers. Cross-fertilization between learning and nonmonotonic reasoning can also occur in such as: the use of answer set solvers for Inductive Logic Programming; speed-up learning while running answer set solvers; learning action theories; learning transition rules in dynamical systems; learning normal, extended and disjunctive programs; formal relationships between learning and nonmonotonic reasoning; abductive learning; updating theories with induction; learning biological networks with inhibition; applications involving default and negation.

This workshop is the first attempt to provide an open forum for the identification of problems and discussion of possible collaborations among researchers with complementary expertise. The workshop was held on September 15th of 2013 in Corunna, Spain. This post-proceedings contains five technical papers (out of six accepted papers) and the abstract of the invited talk by Luc De Raedt.

We would like to thank all the authors who submitted papers, the invited speaker, and all participants who joined discussion at the workshop. We also thank the PC members for their excellent work, as well as the additional reviewers for their contributed to the success of the workshop. Special thanks are due to LPNMR 2013 organizing committee, in particular to Marcello Balduccini for his support as the LPNMR workshop chair.

November 2013

Katsumi Inoue
Chiaki Sakama
Program Chairs

# Organization

## Program Chairs

Katsumi Inoue (National Institute of Informatics, Japan)

Chiaki Sakama (Wakayama University, Japan)

## Program Committee

Dalal Alrajeh (Imperial College London, UK)

Marcello Balduccini (Kodak Research Laboratories, USA)

Chitta Baral (Arizona State University, USA)

Gauvain Bourgne (Universite Pierre et Marie Curie, France)

Luc De Raedt (Katholieke Universiteit Leuven, Belgium)

Francesca A. Lisi (Universita degli Studi di Bari "Aldo Moro", Italy)

Stephen Muggleton (Imperial College London, UK)

Adrian Pearce (University of Melbourne, Australia)

Oliver Ray (University of Bristol, UK)

Taisuke Sato (Tokyo Institute of Technology, Japan)

Torsten Schaub (University of Potsdam, Germany)

## Additional Reviewers

Santiago Videla

# Table of Contents

## Invited Talk

## Regular Papers

## Declarative Modeling for Machine Learning and Data Mining

Luc De Raedt
Katholieke Universiteit Leuven, Belgium

Despite the popularity of machine learning and data mining today, it remains challenging to develop applications and software that incorporates machine learning or data mining techniques. This is because machine learning and data mining have focused on developing high-performance algorithms for solving particular tasks rather than on developing general principles and techniques. I propose to alleviate these problems by applying the constraint programming methodology to machine learning and data mining and to specify machine learning and data mining problems as constraint satisfaction and optimization problems. What is essential is that the user be provided with a way to declaratively specify what the machine learning or data mining problem is rather than having to outline how that solution needs to be computed. This corresponds to a model + solver-based approach to machine learning and data mining, in which the user specifies the problem in a high level modeling language and the system automatically transforms such models into a format that can be used by a solver to efficiently generate a solution. This should be much easier for the user than having to implement or adapt an algorithm that computes a particular solution to a specific problem. Throughout the talk, I shall use illustrations from our work on constraint programming for itemset mining and probabilistic programming.

# Probabilistic Inductive Answer Set Programming by Model Sampling and Counting

Alessandra Mileo[1] and Matthias Nickles[1,2]

[1] Digital Enterprise Research Institute (DERI)
College of Engineering & Informatics
National University of Ireland, Galway
{alessandra.mileo,matthias.nickles}@deri.org
[2] Department of Information Technology
College of Engineering & Informatics
National University of Ireland, Galway

**Abstract.** We propose a new formal language for the expressive representation of probabilistic knowledge based on Answer Set Programming (ASP). It allows for the annotation of first-order formulas as well as ASP rules and facts with probabilities and for learning of such weights from data (parameter estimation). Weights are given a semantics in terms of a probability distribution over answer sets. In contrast to related approaches, we approach inference by optionally utilizing so-called streamlining XOR constraints, in order to reduce the number of computed answer sets. Our approach is prototypically implemented. Examples illustrate the introduced concepts and point at issues and topics for future research.

## 1 Introduction

Reasoning in the presence of uncertainty and dealing with complex relational structures (such as social networks) is required in many prevalent application fields, such as knowledge mining and knowledge representation on the Web, the Semantic Web and Linked Data, due to the high degree of inconsistency and uncertainty of information typically found in these domains. Probabilistic logic programing, and the ability to learn probabilistic logic programs from data, can provide an attractive approach to uncertainty reasoning and relational machine learning, since it combines the deduction power and declarative nature of logic programming with probabilistic inference abilities traditionally known from graphical models such as Bayesian networks. A very successful type of logic programming for nonmonotonic domains is *Answer Set Programming* (ASP) [1, 2]. Since statistical-relational approaches to probabilistic reasoning typically rely heavily on the grounding of first-order or other relational information, and various efficient techniques have been developed to deal with challenging tasks such as weight and structure learning for such grounding-based approaches, ASP looks like an ideal basis for probabilistic logic programming, given its expressiveness and fast and highly developed grounders and solvers. However, despite the successful employment of conceptually related approaches in the area of SAT solving for probabilistic inference tasks, only a small number of approaches to probabilistic knowledge representation or probabilistic inductive logic programming under the stable model semantics exist so far, of which some are rather restrictive wrt. expressiveness and parameter estimation techniques. We build upon these and other existing approaches in the area of probabilistic (inductive) logic programming in order to provide an initial framework consisting of a new ASP-based language (with first-order as well as ASP syntax) for the representation of subjective probabilistic knowledge, and formal and algorithmic tools for inference and parameter estimation from example data. Weights which represent probabilities can be attached to arbitrary formulas, and we show how this can be used to perform probabilistic inference and how further weights can be inductively learned from examples.

The remainder of this paper is organized as follows: the next section presents relevant related approaches. Section 3 introduces syntax and semantics of our new language. Section 4 presents our approach to probabilistic inference, and Section 5 shows how formula weights can be learned from data. Section 6 concludes.

## 2 Related Work

Being one of the early approaches to the logic-based representation of uncertainty sparked by Nilsson's seminal work [3], [4] presents three different probabilistic first-order languages, and compares them with a related approach by Bacchus [5]. One language has a *domain-frequency* (or *statistical*) semantics, one has a possible worlds semantics (like our approach), and one bridges both types of semantics. While those languages as such are mainly of theoretical relevance, their types of semantics still form the backbone of most practically relevant contemporary approaches.
Many newer approaches, including Markov Logic Networks (see below), require a possibly expensive grounding (propositionalization) of first-order theories over finite domains. A recent approach which does not fall into this category but employs the principle of maximum entropy in favor of performing extensive groundings is [6]. However, since ASP is predestined for efficient grounding, we do not see grounding necessarily as a shortcoming. *Stochastic Logic Programs* (SLPs) [7] are an influential approach where sets of rules in form of range-restricted clauses can be labeled with probabilities. Parameter learning for SLPs is approached in [8] using the EM-algorithm. Approaches which combine concepts from Bayesian network theory with relational modeling and learning are, e.g., [9–11]. Probabilistic Relational Models (PRM) [9] can be seen as relational counterparts to Bayesian networks In contrast to those, our approach does not directly relate to graphical models such as Bayesian or Markov Networks but works on arbitrary possible worlds which are generated by ASP solvers. ProbLog [12] allows for probabilistic facts and definite clauses, and approaches to probabilistic rule and parameter learning (from interpretations) also exist for ProbLog. Inference is based on weighted model counting, which is similarly to our approach, but uses Boolean satisfiability instead of stable model search. ProbLog builds upon the very influential Distribution Semantics introduced for PRISM [13], which is also used by other approaches, such as Independent Choice Logic (ICL) [14]. Another important approach outside the area of ASP are *Markov Logic Networks* (MLN) [15], which are related to ours. A MLN consists of first-order formulas annotated with weights (which are not probabilities). MLNs are used as "templates" from which Markov networks are constructed, i.e., graphical models for the joint distribution of a set of random variables.

The (ground) Markov network generated from the MLN then determines a probability distribution over possible worlds. MLNs are syntactically similar to the logic programs in our framework, however, in contrast to MLN, we allow for probabilities as formula weights. Our initial approach to weight learning is closely related to certain approaches to MLN parameter learning (e.g., [16]), as described in Section 5.
Located in the field of nonmonotonic logic programming, our approach is also influenced by P-log [17] and abduction-based rule learning in probabilistic nonmonotonic domains [18]. With P-log, our approaches shares the view that answer sets can be seen as possible worlds in the sense of [3]. However, the syntax of P-log is quite different from our language, by restricting probabilistic annotations to certain syntactical forms and by the concept of independent experiments, which simplifies the implementation of their framework. In distinction from P-log, there is no particular coverage for causality modeling in our framework. [18] allows to associate probabilities with abducibles and to learn both rules and probabilistic weights from given data (in form of literals). In contrast, our present approach does not comprise rule learning. However, our weight learning algorithm allows for learning from any kind of formulas and for the specification of virtually any sort of hypothesis as learning target, not only sets of abducibles. Both [18] and our approach employ gradient descent for weight learning. Other approaches to probabilistic logic programming based on the stable model semantics for the logic aspects include [19] and [20]. [19] appears to be a powerful approach, but restricts probabilistic weighting to certain types of formulas, in order to achieve a low computational reasoning complexity. Its probabilistic annotation scheme is similar to that proposed in [20]. [20] provides both a language and an in-depth investigation of the stable model semantics (in particular the semantics of non-monotonic negation) of probabilistic deductive databases.
Our approach (and ASP in general) is closely related to SAT solving, #SAT and constraint solving. As [21] shows, Bayesian networks can be "translated" into a weighted model counting problem over propositional formulas, which is related to our approach to probabilistic inference, although details are quite different. Also, the XOR constraining approach [22] employed for sampling of answer sets (Section 4) has originally been invented for the sampling of propositional truth assignments.

## 3 Probabilistic Answer Set Programming with PrASP

Before we turn to probabilistic inference and parameter estimation, we introduce our new language for probabilistic non-monotonic logic programming, called Probabilistic Answer Set Programming (*PrASP* ). The main enhancement provided by PrASP compared to definite ASP and related probabilistic approaches to ASP is the possibility to annotate any formulas in first-order syntax (but also AnsProlog rules and facts) with probabilities.

### 3.1 Syntax: Just add probabilities

To remove unnecessary syntax restrictions and because we will later require certain syntactic modifications of given programs which are easier to express in First-Order Logic (FOL) notation, we allow for FOL statements in our logic programs. More precisely, a *PrASP program* consists of ground or non-ground formulas in unrestricted first-order syntax annotated with numerical *weights* (provided by some domain expert or learned from data). Weights directly represent probabilities. If the weights are removed, and provided finite variable domains, any such program can be converted into an equivalent answer set program by means of the transformation described in [23].

Let $\Phi$ be a set of function, predicate and object symbols and $\mathcal{L}(\Phi)$ a first-order language over $\Phi$ and the usual connectives (including both strong negation "-" and default negation "not") and first-order quantifiers.
Formally, a PrASP program is a non-empty finite set $\{([p], f_i)\}$ of PrASP *formulas* where each formula $f_i \in \mathcal{L}(\Phi)$ is annotated with a *weight* $[p]$. A weight directly represents a probability (provided it is probabilistically sound). If the weight is omitted for some formula of the program, weight $[1]$ is assumed. The weight $p$ of $[p]$ $f$ is denoted as $w(f)$.
Let $\Lambda^-$ denote PrASP program $\Lambda$ stripped of all weights. Weights need to be probabilistically sound, in the sense that the system of inequalities (1) - (4) in Section 3.2 must have at least one solution (however, in practice this does not need to be strictly the case, since the constraint solver employed for finding a probability distribution over possible worlds can find approximate solutions often even if the given weights are inconsistent).

In order to translate conjunctions of unweighted formulas in first-order syntax into disjunctive programs with a stable model semantics, we further define transformation $lp : \mathcal{L}(\Phi) \cup dLp(\Phi) \rightarrow dLp(\Phi)$, where $dLp(\Phi)$ is the set of all disjunctive programs over $\Phi$. The details of this transformation can be found in [23][3]. Applied to rules and facts in ASP syntax, $lp$ simply returns these. This allows to make use of the wide range of advanced possibilities offered by contemporary ASP grounders in addition to FOL syntax (such as aggregates), although when defining the semantics of programs, we consider only formulas in FOL syntax.

### 3.2 Semantics

The probabilities attached to formulas in a PrASP program induce a probability distribution over answer sets of an ordinary answer set program which we call the *spanning program* associated with that PrASP program. Informally, the idea is to transform a PrASP program into an answer set program whose answer sets reflect the nondeterminism introduced by the probabilistic weights: each annotated formula might hold as well as not hold (unless its weight is [0] or [1]). Of course, this transformation is lossy, so we need to memorize the weights for the later computation of a probability distribution over possible worlds. The important aspect of the spanning program is that it programmatically generates a set of possible worlds in form of answer sets.
Technically, the spanning program $\rho(\Lambda)$ of PrASP program $\Lambda$ is a disjunctive program obtained by transformation $lp(\Lambda')$. We generate $\Lambda'$ from $\Lambda$ by removing all weights and transforming each formerly weighted formula $f$ into a disjunction $f|not\ f$, where $not$ stands for default negation and $|$ stands for the disjunction in ASP (so probabilities are "default probabilities" in our framework). Note that $f|not\ f$ doesn't guarantee that answer sets are generated for weighted formula $f$. By using ASP choice constructs such as aggregates and disjunctions, the user can basically generate as many answer sets (possible worlds) as desired.

Formulas do not need to be ground - as defined in Section 3.1, they can contain existentially as well as universally quantified variables in the FOL sense (although restricted to finite domains).
As an example, consider the following simple ground[4] PrASP program:

```
[0.7] q <- p.
```

[3] The use of the translation into ASP syntax requires either an ASP solver which can deal directly with disjunctive logic programs (such as claspD) or a grounder which is able to shift disjunctions from the head of the respective rules into the bodies, such as gringo [24].

[4] Examples for PrASP programs with variables are presented later in this paper.

```
[0.3] p.
[0.2] -p & r.
```

The set of answer sets (which we take as possible worlds) of the spanning program of this PrASP program is $\{\{p,q\},\{-p,r\},\{\},\{p\}\}$.

The semantics of a PrASP program $\Lambda$ and single PrASP formulas is defined in terms of a probability distribution over a set of possible worlds (in form of answer sets of $\rho(\Lambda)$) in connection with the stable model semantics. This is analogously to the use of *Type 2 probability structures* [4] for first-order probabilistic logics with subjective probabilities, but restricted to finite domains of discourse.

Let $M = (D, \Theta, \pi, \mu)$ be a probability structure where $D$ is a finite discrete domain of objects, $\Theta$ is a non-empty set of possible worlds, $\pi$ a function which assigns to the symbols in $\Phi$ (see Section 3.1) predicates, functions and objects over/from $D$, and $\mu$ a discrete probability function over $\Theta$.
Each possible world is a Herbrand interpretation over $\Phi$. Since we will use answer sets as possible worlds, defining $\Gamma(a)$ to be the set of all answer sets of answer set program $a$ will become handy. For example, given $\rho(\Lambda)$ as (uncertain) knowledge, the set of worlds deemed possible according to existing belief $\rho(\Lambda)$ is $\Gamma(\rho(\Lambda))$ in our framework.

We define a (non-probabilistic) satisfaction relation of possible worlds and unannotated programs as follows: let $\Lambda^-$ be is an unannotated program. Then $(M,\theta) \models_\Theta \Lambda^-$ *iff* $\theta \in \Gamma(lp(\Lambda^-))$ and $\theta \in \Theta$ (from this it follows that $\Theta$ induces its own closed world assumption - any answer set which is not in $\Theta$ is not satisfiable wrt. $\models_\Theta$). The probability $\mu(\{\theta\})$ of a possible world $\theta$ is denoted as $Pr(\theta)$ and sometimes called "weight" of $\theta$. For a disjunctive program $\psi$, we analogously define $(M,\theta) \models_\Theta \psi$ *iff* $\theta \in \Gamma(\psi)$ and $\theta \in \Theta$.

To do groundwork for the computation of a probability distribution over possible worlds $\Theta$ which are "generated" and weighted by some given background knowledge in form of a PrASP program, we define a (non-probabilistic) satisfaction relation of possible worlds and unannotated formulas: let $\phi$ be a PrASP formula (without weight) and $\theta$ be a possible world. Then $(M,\theta) \models_\Lambda \phi$ *iff* $(M,\theta) \models_\Theta \rho(\Lambda) \cup lp(\phi)$ *and* $\Theta = \Gamma(\rho(\Lambda))$ (we say formula $\phi$ is *true in possible world* $\theta$). Sometimes we will just write $\theta \models_\Lambda \phi$ if $M$ is given by the context. A notable property of this definition is that it does not restrict us to single ground formulas. Essentially, an unannotated formula $\phi$ can be any answer set program specified in FOL syntax, even if its grounding consists of multiple sentences. Observe that $\Theta$ restricts $\models_\Lambda$ to answer sets of $\rho(\Lambda)$. For convenience, we will abbreviate $(M,\theta) \models_\Lambda \phi$ as $\theta \models_\Lambda \phi$.

$Pr(\phi)$ denotes the probability of a formula $\phi$, with $Pr(\phi) = \mu(\{\theta \in \Theta : (M,\theta) \models_\Lambda \phi\})$. Note that this holds both for annotated and unannotated formulas: even if it has a weight attached, the probability of a PrASP formula is defined by means of $\mu$ and only indirectly by its manually assigned weight (weights are used below as constraints for the computation of a probabilistically consistent $\mu$). Further observe that there is no particular treatment for conditional probabilities in our framework; $Pr(a|b)$ is simply calculated as $Pr(a \wedge b)/Pr(b)$.
While our framework so far is general enough to account for probabilistic inference using unrestricted programs and query formulas (provided we are given a probability distribution over the possible answer sets), this generality also means a high complexity in terms of computability for inference-heavy tasks which rely on the repeated application of operator $\models_\Lambda$, even if we would avoid the transformation $lp$ and restrict ourselves to the use of ASP syntax. However, for many formulas $\phi$, in particular query formulas (i.e., formulas whose probability or truth we would like to infer), we can avoid costly calls of some ASP solver for the computation of $\theta \models_\Lambda \phi$ (given $\theta$), by transforming the formula into propositional syntax (which is often trivial) and computing its disjunctive normal form (DNF). While conversion into DNF takes exponential time in the worst case, for small formulas which are already close (or even identical) to their DNF it can be very efficient. Since a possible world is a Herbrand interpretation in form of a set of literals (such as an answer set), $\phi$ is true in possible world $\theta$ if any clause of the DNF of $\phi$ is a subset of $\theta$ (occurrence of default negation in literals requires a special handling, by treating $not\ atom$ in a DNF clause as a test for absence of $atom$ in $\theta$).

The obvious question now, addressed before for other probabilistic logics, is how to compute $\mu$, i.e., how to obtain a probability distribution over possible worlds (which tells us for each possible world the probability with which this possible world is the actual world) from a given annotated program $\Lambda$ in a sound and computationally inexpensive way.

Generally, we can express the search for probability distributions in form of a number of constraints which constitute a system of linear inequalities (which reduce to linear equalities for point probabilities as weights). This system typically has multiple or even infinitely many solutions (even though we do not allow for probability intervals) and computation can be costly, depending on the number of possible worlds according to $\rho(\Lambda)$.

We define the parameterized probability distribution $\mu(\Lambda, \Theta)$ over a set $\Theta$ of answer sets as the solution (for all $Pr(\theta_i)$) of the following system of linear equations and an inequality (if precisely one solution exists) or as the solution with maximum entropy [6], in case multiple solutions exist [5]. We require that the given weights in a PrASP program are chosen such that the following constraint system has at least one solution.

$$\sum_{\theta_i \in \Theta : \theta_i \models_\Lambda f_1} Pr(\theta_i) = w(f_1) \quad (1)$$

$$\cdots$$

$$\sum_{\theta_i \in \Theta : \theta_i \models_\Lambda f_n} Pr(\theta_i) = w(f_n) \quad (2)$$

$$\sum_{\theta_i \in \Theta} \theta_i = 1 \quad (3)$$

$$\forall \theta_i \in \Theta : 0 \leq Pr(\theta_i) \leq 1 \quad (4)$$

At this, $\Lambda = \{f_1, ..., f_n\}$ is a PrASP program.

The *canonical probability distribution* $\mu(\Lambda)$ of $\Lambda$ is defined as $\mu(\Lambda, \Gamma(\rho(\Lambda)))$. In the rest of the paper, we refer to $\mu(\Lambda)$ when we refer to the probability distribution over the answer sets of the spanning program of a given PrASP program $\Lambda$.

## 4 Inference

Given possible world weights ($\mu(\Lambda)$), probabilistic inference becomes a model counting task where each model has a weight: we can compute the probability of any query formula $\phi$ by summing up the probabilities (weights) of those possible worlds (models) where $\phi$ is true. To make this viable even for larger sets of possible worlds, we optionally restrict the calculation of $\mu(\Lambda)$ to a number of answer sets sampled near-uniformly at random from the total set of answer sets of the spanning program, as described in Section 4.1.

[5] Since in this case the number of solutions of the system of linear equations is infinite, de facto we need to choose the maximum entropy solution of some finite subset. In the current prototype implementation, we generate a user-defined number of random solutions derived from a solution computed using a constrained variant of Singular Value Decomposition and the null space of the coefficient matrix of the system of linear equations (1)-(3).

### 4.1 Adding a sampling step and computing probabilities

All tasks described so far (solving the system of (in)equalities, counting of weighted answer sets) become intractable for very large sets of possible worlds. To tackle this issue, we want to restrict the application of these tasks to a sampled subset of all possible worlds. Concretely, we want to find a way to sample (near-)uniformly from the total set of answer sets *without* computing a very large number of answer sets. While this way the set of answer sets cannot be computed using only a single call of the ASP solver but requires a number of separate calls (each with different sampling constraints), the required solver calls can be performed *in parallel*. However, a shortcoming of the sampling approach is that there is currently no way to pre-compute the size of the minimally required set of samples.

Guaranteeing near-uniformity in answer set sampling looks like a highly non-trivial task, since the set of answers obtained from ASP solvers is typically not uniformly distributed but strongly biased in hardly foreseeable ways (due to various interplaying heuristics applied by modern solvers), so we could not simply request any single answer set from the solver.

However, we can make use of so-called *XOR constraints* (a form of streamlining constraints in the area of SAT solving) for near-uniform sampling [22] to obtain samples from the space of all answer sets, within arbitrarily narrow probabilistic bounds, using any off-the-shelf ASP solver. Compared to approaches which use Markov Chain Monte Carlo (MCMC) methods to sample from some given distribution, this method has the advantage that the sampling process is typically faster and that it requires only an off-the-shelf ASP solver (which is in the ideal case employed only once per sample, in order to obtain a single answer set). However, a shortcoming is that we are not doing Importance Sampling this way - the probability of a possible world is not taken into account but computed later from the samples.

Counting answer sets could also be achieved using XOR constraints, however, this is not covered in this paper, since it does not comprise weighted counting, and we could normally not use an unweighted counting approach directly.

XOR constraints were originally defined over a set of propositional variables, which we identify with a set of ground atoms $V = \{a_1, ..., a_n\}$. Each XOR constraint is represented by a subset $D$ of $V \cup \{true\}$. $D$ is satisfied by some model if an *odd* number of elements of $D$ are satisfied by this model (i.e., the constraint acts like a parity of $D$). In ASP syntax, an XOR constraint can be represented for example as `:- #even{` $a_1$`,` `...,` $a_n$ $\}$ [25].

In our approach, XOR constraints are independently at random drawn from a probability distribution $\mathbb{X}(|V|, 0.5)$ over the set $V$ of all possible XOR constraints over all ground atoms of the ground answer set program resulting from $\rho(\Lambda)$. $\mathbb{X}(|V|, 0.5)$ is defined such that each XOR constraint is drawn from this distribution independently at

random with probability 0.5 and includes $true$ with probability 0.5. In effect, any given XOR constraint is drawn with probability $2^{-(|V|+1|)}$ (see [22] for details). Since adding an XOR constraint to an answer set program eliminates any given answer set with probability 0.5, it cuts the set of answer sets in half in expectation. Iteratively adding a small number of XOR constraints to an answer set program therefore reduces the number of answer sets to a small number also. If this process results in a single answer set, the remaining answer set is drawn near-uniformly from the original set of answer sets, as shown in [22].
Since for answer set programs the costs of repeating the addition of constraints until precisely a single answer set remains appears to be higher than the costs of computing somewhat too many models, we just estimate the number of required constraints and choose randomly from the resulting set of answer sets. The following way of answer set sampling using XOR constraints has been used before in Xorro (a tool which is part of the *Potassco* set of ASP tools [25]) in a very similar way.

***sample***: $\psi \mapsto \gamma$

Given any disjunctive program $\psi$, the following procedure computes a random sample $\gamma$ from the set of all answer sets of $\psi$:
$\psi_g \leftarrow$ ground($\psi$)
$ga \leftarrow atoms(\psi_g)$
$xors \leftarrow$ XOR constraints $\{xor_1, ..., xor_n\}$ over $ga$, drawn from $\mathbb{X}(|V|, 0.5)$
$\psi' \leftarrow \psi \cup xors$
$\gamma \leftarrow$ an answer set selected randomly from $\Gamma(\psi')$

At this, the number of constraints $n$ is set to a value large enough to produce one or a very low number of answer sets ($log_2(|ga|)$ in our experiments).
We can now compute $\mu(\Lambda, \Theta')$ (i.e., $Pr(\theta)$ for each $\theta \in \Theta'$) for a set of samples $\Theta'$ obtained by multiple (ideally parallel) calls of *sample* from the spanning program $\rho(\Lambda)$ of PrASP program $\Lambda$, and subsequently sum up the weights of those samples (possible worlds) where the respective query formula (whose marginal probability we want to compute) is true. Precisely, we approximate $Pr(\phi)$ for a (ground or non-ground) query formula $\phi$ using:

$$Pr(\phi) \approx \sum_{\{\theta' \in \Theta' : \theta' \models_\Lambda \phi\}} Pr(\theta') \tag{5}$$

for a sufficiently large set $\Theta'$ of samples. $\theta' \models_\Lambda \phi$ might be calculated using the alternative DNF approach (see Section 3.2), since otherwise any performance gain through sampling might become void.
Conditional probabilities $Pr(a|b)$ can simply be computed as $Pr(a \wedge b)/Pr(b)$.

If no sampling is required (i.e., if the total number of answer sets $\Theta$ is expected to be moderate), inference is done in the same way, we just set $\Theta' = \Theta$.

As an example for inference using our current implementation, consider the following PrASP formalization of a simple coin game:

```
coin(1..3).
[0.6] coin_out(1,heads).
[[0.5]] coin_out(N,heads) :- coin(N), N != 1.
1{coin_out(N,heads), coin_out(N,tails)}1 :- coin(N).
n_win :- coin_out(N,tails), coin(N).
win :- not n_win.
```

At this, the line starting with `[[0.5]]...` is syntactic sugar for a set of weighted rules where variable N is instantiated with all its possible values (i.e.,

`[0.5] coin_out(2,heads) :- coin(2), 2 != 1` and `[0.5] coin_out(3,heads) :- coin(3), 3 != 1`). It would also be possible to use `[0.5]` as annotation of this rule, in which case the weight 0.5 would specify the probability of the whole non-ground formula instead.
Our prototypical implementation accepts query formulas in format `[?] a` (computes the marginal probability of a) and `[?|b] a` (computes the conditional probability $Pr(a|b)$). E.g.,

```
[?] coin_out(1,tails).
[?] coin_out(1,heads) | coin_out(1,tails).
[?] coin_out(1,heads) & coin_out(2,heads) & coin_out(3,heads).
[?] win.
[?|coin_out(1,heads) & coin_out(2,heads) & coin_out(3,heads)] win.
```

...yields the following result

```
[0.3999999999999999] coin_out(1,tails).
[1] coin_out(1,heads) | coin_out(1,tails).
[0.15] coin_out(1,heads) & coin_out(2,heads) & coin_out(3,heads).
[0.15] win.
[1|coin_out(1,heads) & coin_out(2,heads) & coin_out(3,heads)] win.
```

This example also demonstrates that FOL and logic programming / ASP syntax can be freely mixed in background knowledge and queries.
In this example, use of sampling does not make any difference due to its small size. An example where a difference can be observed is presented in Section 5.

## 5 Weight Learning

The general task of parameter learning in probabilistic inductive logic programming is to find probabilistic parameters (weights) of logical formulas which maximize the likelihood given some data (learning examples) [26]. In our case, the hypothesis $H$ (a set of formulas without weights) is provided by an expert, optionally together with some PrASP program as background knowledge $B$. The goal is then to discover weights $w$ of the formulas $H$ such that $Pr(E|H_w \cup B)$ is maximized given example formulas $E = e_1, e_2, ...$. Formally, we want to compute

$$argmax_w(Pr(E|H_w \cup B)) = argmax_w(\prod_{e_i \in E} Pr(e_i|H_w \cup B)) \quad (6)$$

(Making the usual i.i.d. assumption regarding the individual examples in $E$. $H_w$ denotes the hypothesis weighted with weight vector $w$.)

This results in an optimization task which is related but not identical to weight learning for, e.g., Markov Logic Networks (MLNs) and [18]. In MLNs, typically a database (possible world) is given whose likelihood should be maximized, e.g. using a generative approach [16] by gradient descent. Another related approach distinguishes a priori between evidence atoms $X$ and query atoms $Y$ and seeks to maximize the likelihood $Pr(Y|X)$, again using gradient descent [27]. At this, cost-heavy inference is avoided as far as possible, e.g., by optimization of the pseudo-(log-)likelihood instead ot the (log-)likelihood or by approximations of costly counts of true formula groundings in a certain possible world (the basic computation in MLN inference). In contrast, the current implementation of PrASP learns weights from any formulas and not just literals (or, more precisely as for MLNs: atoms, where negation is implicit using a closed-world assumption). Furthermore, the maximization targets are different ($Pr(possible\ world)$ or $Pr(Y|X)$) vs. $Pr(E|H_w \cup B)$).

Regarding the need to reduce inference when learning, PrASP parameter estimation should in principle make no exception, since inference can still be costly even when probabilities are inferred only approximately by use of sampling. However, in our preliminary experiments we found that at least in relatively simple scenarios, there is no need to resort to inference-free approximations such as pseudo-(log-)likelihood. The pseudo-(log-)likelihood approach presented in early works on MLNs [28] would also require a probabilistic ground formula independence analysis in our case, since in PrASP there is no obvious equivalent to Markov blankets.
Note that we assume that the example data is non-probabilistic and fully observable.

Let $H = \{f_1, ..., f_n\}$ be a given set of formulas and a vector $w = (w^1, ..., w^n)$ of (unknown) weights of these formulas. Using the Barzilai and Borwein method [29] (a variant of the gradient descent approach with possibly superlinear convergence), we seek to find $w$ such that $Pr(E|H_w \cup B)$ is maximized ($H_w$ denotes the formulas in $H$ with the weights $w$ such that each $f_i$ is weighted with $w^i$). Any existing weights of formulas in the background knowledge ar not touched, which can significantly reduce learning complexity if $H$ is comparatively small. Probabilistic or unobservable examples are not considered.
The learning algorithm [29] is as follows:

Repeat for $k = 0, 1, ...$ until convergence:
  Set $s_k = \frac{1}{\alpha_k} \nabla(Pr(E|H_{w_k} \cup B))$
  Set $w_{k+1} = w_k + s_k$
  Set $y_k = \nabla(Pr(E|H_{w_{k+1}} \cup B)) - \nabla(Pr(E|H_{w_k} \cup B))$
  Set $\alpha_{k+1} = \frac{s_k^T y_k}{s_k^T s_k}$

At this, the initial gradient ascent step size $\alpha_0$ and the initial weight vector $w_0$ can be chosen freely. $Pr(E|H_w \cup B)$ denotes $\prod_{e_i \in E} Pr(e_i|H_w \cup B)$ inferred using vector $w$ as weights for the hypothesis formulas, and

$$\bigtriangledown(Pr(E|H_w \cup B)) = (\frac{\partial}{\partial w^1} Pr(E|H_w \cup B), ..., \frac{\partial}{\partial w^n} Pr(E|H_w \cup B)) \quad (7)$$

Since we usually cannot practically express $Pr(E|H_w \cup B)$ in dependency of $w$ in closed form, at a first glance, the above formalization appears to be not very helpful. However, we can still resort to numerical differentiation and approximate

$$\bigtriangledown(Pr(E|H_w \cup B)) = \quad (8)$$

$$(\lim_{h \to 0} \frac{Pr(E|H_{(w^1+h,...,w^n)} \cup B) - Pr(E|H_{(w^1,...,w^n)} \cup B)}{h}, \quad (9)$$

$$...,$$

$$\lim_{h \to 0} \frac{Pr(E|H_{(w^1,...,w^n+h)} \cup B) - Pr(E|H_{(w^1,...,w^n)} \cup B)}{h}) \quad (10)$$

by computing the above vector (dropping the limit operator) for a sufficiently small $h$ (in our prototypical implementation, $h = \sqrt{\epsilon} w_i$ is used, where $\epsilon$ is an upper bound to the rounding error using the machine's double-precision floating point arithmetic).
This approach has the benefit of allowing in principle for any maximization target (not just $E$). In particular, any unweighted formulas (unnegated and negated facts as well as rules) can be used as (positive) examples.

As a small example both for inference and weight learning using our preliminary implementation, consider the following fragment of a an indoor localization scenario, which consists of estimating the position of a person, and determining how this person moves a certain number of steps around the environment until a safe position is reached:

```
[0.6] moved(1).
[0.2] moved(2).
point(1..100).
1{atpoint(X):point(X)}1.
distance(1) :- moved(1).
distance(2) :- moved(2).
atpoint(29) | atpoint(30) | atpoint(31) | atpoint(32) | atpoint(33)
 | atpoint(34) | atpoint(35) | atpoint(36) | atpoint(37) -> selected.
safe :- selected, not exception.
exception :- distance(1).
```

The spanning program of this example has 400 answer sets. Inference of $Pr(safe|distance(2))$ and $Pr(safe|distance(1))$ without sampling, but using the DNF approach to computing $\models_{\Lambda}$ (see Section 3.2), requires ca. 1534 ms using our current unoptimized prototype implementation. Using 50 near-uniformly drawn samples, time spent on inference becomes about 936 ms (again using the DNF approach). This time could be further reduced by integrating an ASP solver directly into the inference implementation (a technical drawback of our current implementation is the fact that for each sample, an external ASP solver needs to be called, which causes a significant time overhead, even though these calls can be performed partially in parallel). To demonstrate how the probability of a certain hypothesis can be learned in this scenario, we remove `[0.6] moved(1)` from the program above and turn this formula (without the weight annotation) into a hypothesis. Given example data `safe`, parameter estimation results in $Pr(moved(1)) \approx 0$, learned in 4955 ms without sampling, for 100 points. Using 50 samples, learning time is only 2752 ms. However, the optimal number of samples currently needs to be manually determined (by trying). There is currently no approach yet to find automatically a size of the set of samples such that learning (or inference) time is lower than without sampling without generating unacceptably inaccurate results (also see Section 6).

## 6 Conclusions

With this paper, we have presented a novel framework for uncertainty reasoning and parameter estimation based on Answer Set Programming, with support for probabilistically weighted first-order formulas in background knowledge, hypotheses and queries. While our current framework certainly leaves room for future improvements, we believe that we have already pointed out a new venue towards more practicable probabilistic inductive answer set programming with a high degree of expressiveness. Future work should focus on theoretical analysis (in particular regarding minimum number of samples wrt. inference accuracy), empirical evaluation and on the investigation of viable approaches to PrASP structure learning.

## References

1. Lifschitz, V.: Answer set programming and plan generation. AI **138** (2002) 39–54
2. Gelfond, M., Lifschitz, V.: The stable model semantics for logic programming. In: Proc. of the 5th Int'l Conference on Logic Programming. Volume 161. (1988)
3. Nilsson, N.J.: Probabilistic logic. Artificial Intelligence **28(1)** (1986) 71–87
4. Halpern, J.Y.: An analysis of first-order logics of probability. Artificial Intelligence **46** (1990) 311–350
5. Bacchus, F.: $l_p$, a logic for representing and reasoning with statistical knowledge. Computational Intelligence **6** (1990) 209–231

6. Thimm, M., Kern-Isberner, G.: On probabilistic inference in relational conditional logics. Logic Journal of the IGPL **20** (2012) 872–908
7. Muggleton, S.: Learning stochastic logic programs. Electron. Trans. Artif. Intell. **4** (2000) 141–153
8. Cussens, J.: Parameter estimation in stochastic logic programs. In: Machine Learning. (2000) 2001
9. Friedman, N., Getoor, L., Koller, D., Pfeffer, A.: Learning probabilistic relational models. In: In IJCAI, Springer-Verlag (1999) 1300–1309
10. Kersting, K., Raedt, L.D.: Bayesian logic programs. In: Proceedings of the 10th International Conference on Inductive Logic Programming. (2000)
11. Laskey, K.B., Costa, P.C.: Of klingons and starships: Bayesian logic for the 23rd century. In: Proceedings of the Twenty-first Conference on Uncertainty in Artificial Intelligence. (2005)
12. Raedt, L.D., Kimmig, A., Toivonen, H.: Problog: A probabilistic prolog and its application in link discovery. In: IJCAI. (2007) 2462–2467
13. Sato, T., Kameya, Y.: Prism: a language for symbolic-statistical modeling. In: In Proceedings of the 15th International Joint Conference on Artificial Intelligence (IJCAI97. (1997) 1330–1335
14. Poole, D.: The independent choice logic for modelling multiple agents under uncertainty. Artificial Intelligence **94** (1997) 7–56
15. Richardson, M., Domingos, P.: Markov logic networks. Mach. Learn. **62** (2006) 107–136
16. Lowd, D., Domingos, P.: Efficient weight learning for markov logic networks. In: In Proceedings of the Eleventh European Conference on Principles and Practice of Knowledge Discovery in Databases. (2007) 200–211
17. Baral, C., Gelfond, M., Rushton, J.N.: Probabilistic reasoning with answer sets. CoRR vol. abs/0812.0659 (2008)
18. Corapi, D., Sykes, D., Inoue, K., Russo, A.: Probabilistic rule learning in nonmonotonic domains. In: Proceedings of the 12th international conference on Computational logic in multi-agent systems. CLIMA'11, Berlin, Heidelberg, Springer-Verlag (2011) 243–258
19. Saad, E., Pontelli, E.: Hybrid probabilistic logic programming with non-monotoic negation. In: In Twenty First International Conference on Logic Programming, Springer Verlag (2005)
20. Ng, R.T., Subrahmanian, V.S.: Stable semantics for probabilistic deductive databases. Inf. Comput. **110** (1994) 42–83
21. Sang, T., Beame, P., Kautz, H.A.: Performing bayesian inference by weighted model counting. In: AAAI. (2005) 475–482
22. Gomes, C.P., Sabharwal, A., Selman, B.: Near-uniform sampling of combinatorial spaces using xor constraints. In: NIPS. (2006) 481–488
23. Lee, J., Palla, R.: System f2lp - computing answer sets of first-order formulas. In Erdem, E., Lin, F., Schaub, T., eds.: LPNMR. Volume 5753 of Lecture Notes in Computer Science., Springer (2009) 515–521
24. Gebser, M., Kaufmann, B., Schaub, T.: Conflict-driven answer set solving: From theory to practice. Artificial Intelligence (2012)
25. Gebser, M., Kaufmann, B., Kaminski, R., Ostrowski, M., Schaub, T., Schneider, M.: Potassco: The potsdam answer set solving collection. AI Commun. **24** (2011) 107–124
26. Raedt, L.D., Kersting, K.: Probabilistic inductive logic programming. In: Probabilistic Inductive Logic Programming. (2008) 1–27
27. Huynh, T.N., Mooney, R.J.: Discriminative structure and parameter learning for markov logic networks. In: 25th Int. Conf. on. (2008) 416–423
28. Richardson, M., Domingos, P.: Markov logic networks. Machine Learning **62** (2006) 107–136
29. Barzilai, J., Borwein, J.M.: Two point step size gradient methods. IMA J. Numer. Anal. (1988)

# Inference and Learning of Boolean Networks using Answer Set Programming

Alexandre Rocca, Tony Ribeiro, and Katsumi Inoue

National Institute of Informatics.
2-1-2 Hitotsubashi, Chiyoda-ku, Tokyo 101-8430, Japan

**Abstract.** A *Boolean Network* is a compact mathematical representation of biological systems widely used in bioinformatics. However, in practice, experiments are usually not sufficient to infer a Boolean network which represents the whole biological system. Previous works relied on *inferring and learning* techniques to complete those models, or to learn new networks satisfying experimental properties represented as temporal logic properties. In this work, we use the *Answer Set Programming* (ASP), a highly expressive declarative language with fast solvers, to provide an efficient, and easily adaptable approach to learn/complete Boolean networks. We use the fast generation-constraint approach of the ASP, with temporal logic specifications, to learn and infer a minimal transition model of a Boolean network.

## 1 Introduction

A *Boolean Network* (BN) is a compact mathematical representation widely used in bioinformatics [13–15]. Initially introduced to represent gene regulatory networks by [13], Boolean Networks have been used in many research fields to represent other Boolean interaction system such as electronic circuits [4] or social interaction models [10]. In recent years, there is a growing interest in the development of techniques for analysis and learning of Boolean networks.

Some works like [7], focus on finding cycle, i.e. attractors, in the behaviour of the system. Detecting attractors and their basins of attraction are very important for analysts to ensure non-time dependant property of a system. In the case of electronic circuits, analysis techniques can also be used to perform *model checking*: ensure that the system behaviour is correct. Some other works develop methods to construct a BN. In [12], the authors proposed a framework to learn the dynamics of a BN from the interpretation of its states transitions, but not from general expression like with temporal logics. In bioinformatics, learning the dynamics of a biological systems helps in identifying the influence of genes and designing more efficient drugs.

In this paper, we propose a model checking framework based on Answer Set Programming dedicated to Boolean Network. Answer set programming (ASP) [9] is a form of declarative programming that has been successfully used in many model-checking problems [11, 1, 17]. This framework allows us to check temporal logic properties against a Boolean Network represented in ASP. The temporal logic is an extension of the propositional logic that can describe properties on dynamical behaviours. In particular, we provide an ASP translation of the Linear Time Logic (LTL) [16] and the Computational Tree Logic (CTL) [6]. To check those properties, we use well known *model checking* techniques similar to the ones used in the model checkers of [2, 5].

The novelty of our model checking framework is the possibility to analyse and infer Boolean Network using ASP. Many model checkers have already been proposed for the analysis of Boolean network. The most similar to our framework are [2], and [5]. However, these model checker rely on SAT and/or BDD approaches to solve the problem. Like [3], our framework can complete an existing Boolean Network by ensuring temporal logic properties; where other work like [15] (focusing on the Consistency Problem), complete or learn a BN by satisfying experimental properties. But, again, our framework use ASP whereas the approach in [3] uses SAT and BDD approaches. ASP takes advantage of the expressiveness of first order logic and high performance solvers like *clasp* [8] make it an interesting alternative to SAT/BDD-based approaches[1]. If there are some previous work about model-checking using ASP, to the best of our knowledge, none of them consider both analysis, construction and completion of Boolean Networks over LTL/CTL properties.

## 2 Preliminary

### 2.1 Boolean Network

A Boolean network (BN) [13] is a pair (N,F) with N = $\{n_1, \ldots, n_k\}$ a finite set of nodes and F = $\{f_1, \ldots, f_k\}$ a corresponding set of Boolean functions. In the case of a gene regulatory network, nodes represent genes and Boolean function represent their relations. If $n_i$(t) represents the value of $n_i$ at the time t of computation, then $n_i$ takes either 1 (expressed) or 0 (not expressed). A vector (or **state**) $s(t) = (n_1(t), \ldots, n_k(t))$ is the expression of the nodes in N at time step t. There are $2^k$ possible states for each time step. The state of a node $n_i$ at the next time step $t+1$ is determined by $n_i(t+1)$=$f_i(n_{i_1}(t), \ldots ,n_{i_p}(t))$, with $n_{i_1}, \ldots ,n_{i_p}$ the nodes directly influencing $n_i$, and also called regulation nodes of $n_i$. Boolean networks can be represented by three different ways: the interaction graph (see Fig. 1), the written diagram which represents the transitions between $n_i(t)$ and $n_i(t+1)$, and the truth table. From the truth table we can create the state-transitions diagram. The value of nodes can be updated synchronously, or asynchronously. A *Synchronous Boolean network* (SBN) is a network where all the nodes are updated at the same time. The successive sequence of states during an execution, called trajectory of a BN, or path, is deterministic in a SBN. An *Asynchronous Boolean network* (ABN) is a network where one node may be updated at given time time. A ABN path can be non deterministic.
One of the interesting properties of the Boolean network is the attractors. Given a set S=($s_1$,...,$s_n$), and a reachability function R. Then R($s_i$) are the reachable states from any path starting from $s_i$. Then, S is an attractor if for any state $s_i$ ∈ S, R($s_i$)=S. Attractors represent the stable states of a Boolean network, and describe a stability in the behaviour.

[1] If we compare with the work on qualitative models

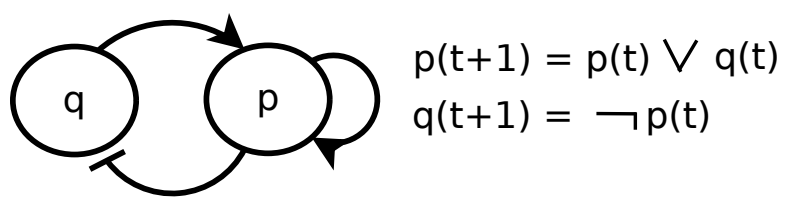

**Fig. 1.** Example of Boolean network

### 2.2 Temporal Logic

In model checking, a model is described by a Kripke structure. A Kripke structure is $M(S, I, T, L)$, with S a set of states, I⊆S a set of initial states, T⊆ $S \times S$ the transition relations, and L: S→P(A) a labelling function, with A the set of atomic propositions and P(A) the powerset of A. For each state $s$ ∈S, L(s) is the set of atomic propositions which are true in s. The behaviour of M is defined by paths. A path $p$ of $M$ is a succession of states ($s_0$,$s_1$,...), where $s_i$ ∈S and T($s_i$,$s_{i+1}$) holds for all $i \geq 0$. The i-th state of a path is denoted $p(i)$.
The temporal logic is an extension of the propositional logic, to describe properties of a system. First the *Linear Temporal Logic* (LTL) is defined as follow:

$$\varphi ::= \text{a} \in \text{A} | \neg\varphi | \varphi_1 \wedge \varphi_2 | \varphi_1 \vee \varphi_2 | G\varphi | \varphi_1 U \varphi_2 | X\varphi | F\varphi | \varphi_1 R \varphi_2 | \Rightarrow$$

$p \models a$ iff $a \in L(p(0))$ $\quad$ $p \models \neg\varphi$ iff $p \nvDash \varphi$
$p \models \varphi_1 \wedge \varphi_2$ iff $p \models \varphi_1$ and $p \models \varphi_2$ $\quad$ $p \models \varphi_1 \vee \varphi_2$ iff $p \models \varphi_1$ or $p \models \varphi_2$
$p \models G\varphi$ iff $p(i) \models \varphi$ ∀i ≥0 $\quad$ $p \models \text{X}\varphi$ iff $p(1) \models \varphi$

$p \models \varphi_1 \text{U} \varphi_2$ iff ∃i≥0 | $p(i) \models \varphi_2$ and ∀0≤ $k \leq i$ $p(k) \models \varphi_1$

From these formulas, we can build any other LTL formulas:
$p \models \varphi_1 \Rightarrow \varphi_2$ iff $p \models \neg(\varphi_1 \wedge \neg\varphi_2)$
$p \models \text{F}\varphi$ iff $p \models \top\text{U}\varphi$, and $p \models \varphi_1 \text{R} \varphi_2$ iff $p \models \neg\varphi_1 \text{U} \neg\varphi_2$
We note that verifying a property on a given path $p$ is equivalent to verifying the property on the initial state of the path.

The computational Tree Logic (CTL) is an extension of propositional logic to describe properties on a branching time behaviour. Like in the LTL description, we use a Kripke model to describe the system. We can separate the CTL operators in two classes: the Global operators with a $A$, and the Existential operators with a $E$. If the LTL describes properties on paths, CTL does it on set of path. The CTL syntax is the following:

$$\varphi ::= a \in \mathrm{A} | \Rightarrow | \neg\varphi | \varphi_1 \wedge \varphi_2 | \varphi_1 \vee \varphi_2 | EG\varphi |$$
$$E\varphi_1 U \varphi_2 | AX\varphi | EF\varphi | AG\varphi | A\varphi_1 U \varphi_2 | AX\varphi | AF\varphi$$

For the description of the properties, the common part with the LTL ($p$,¬,∧,∨) will not be explicated again. For $M = (S, I, T, L)$ a Kripke model and $s \in$S:

(M,s) $\models$ EG$\varphi$ iff ∃ a path $p$ | $p(0) = s$ and ∀ 0≤i (M,$s_i$=$p(i)$)$\models \varphi$
(M,s) $\models$ E$\varphi_1$U$\varphi_2$ iff ∃ a path $p$ | p(0)=s and ∃i≥0 | (M,$s_i$=p(i))$\models$ $\varphi_2$ and ∀0≤k≤i (M,$s_k$=p(k))$\models \varphi_1$
(M,s) $\models$ EX$\varphi$ iff ∃ a path $p$ | $p(0) = s$ and (M,$s_1$=$p(1)$)$\models \varphi$

Same as before, the other CTL formulas can be defined from those three:

(M,s) $\models$ AG$\varphi$ iff (M,s) $\models$ ¬EF¬$\varphi$
(M,s) $\models$ A$\varphi_1$U$\varphi_2$ iff (M,s) $\models$ ¬(E(¬$\varphi_1$U(¬$\varphi_1 \wedge \varphi_2$)) ∧ ¬EG(¬$\varphi_2$))
(M,s) $\models$ AX$\varphi$ iff (M,s) $\models$ (M,s) $\models$ ¬EX¬$\varphi$
(M,s) $\models$ AF$\varphi$ iff (M,s) $\models$ (M,s) $\models$ ¬EG¬$\varphi$

## 3 Inferring a non complete Boolean network

Boolean networks constructed from real life observations are often incomplete, especially in biology: there is often interactions between two genes (represented by two Boolean nodes) that are unknown, or ambiguous.
In this section, we first focus on how to complete a Boolean network thanks to some experimental data expressed as temporal logic formulas. The Boolean network given as input can be synchronous, or asynchronous, and the temporal logics used will be the CTL and the LTL.
The number of possible completed network of an incomplete Boolean network correspond to the number of possible behaviours of this network. There is at most $n$ ambiguous interactions (if there is $n$ nodes) per node and each ambiguous interactions can be either an activation, an inhibition or with no effect so that in the worst case there is $n3^n$ possible behaviours. In the first BN of Figure 2, the influence of $x_2$ on $x_3$ is unknown so that there should be $3 * 3^1 = 9$ possibilities. But here the influence of $x_1$ on $x_3$ and $x_3$ among itself is partially known and the number of possibilities is only 5. In the second BN of Figure 3, we know that $x_2$ has an influence on $x_3$ according to $x_1$, so that the number of possibilities is only 3. We can either choose to complete the interaction graph, or the boolean functions, in both case the reasoning is the same.

In the following sections, we propose a method which combines ASP with model-checking techniques to compute the complete models of an ambiguous

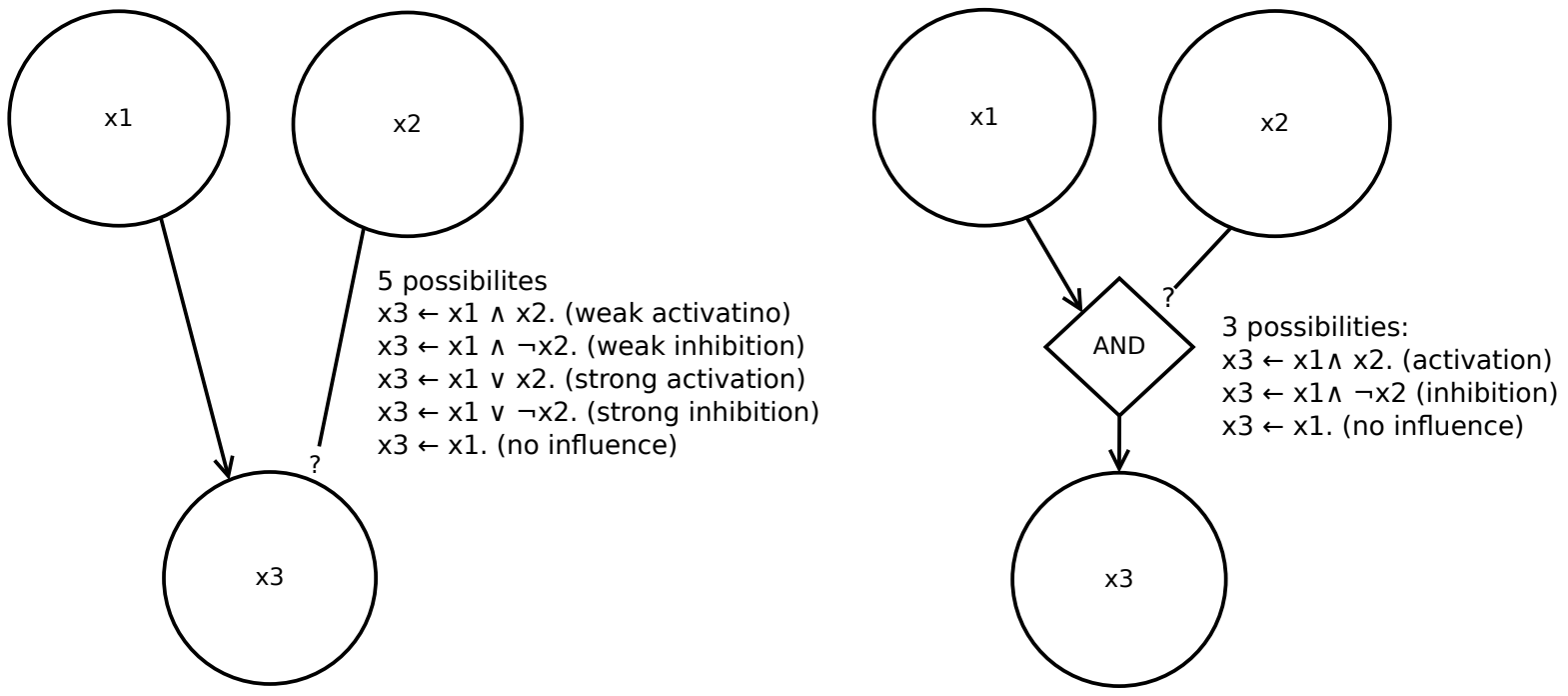

**Fig. 2.** Example of incomplete Boolean network

BN, keeping only the ones with a behaviour satisfying a set of LTL, or CTL, formulas. The techniques we use can be divided in two part: bounded and non-bounded model-checking. The bounded model-checking consists on finite computations: the temporal logic formulas are checked for a limited number of steps and/or a limited run time. For the non-bounded model-checking, the temporal logic formulas are checked for a potentially infinite computation. The following, describes both techniques and their use for the inference: starting by the LTL (Section. 3.1), followed by the CTL (Section. 3.2).

### 3.1 Inference and LTL model-checking in ASP

LTL model-checking verifies properties on a linear path. This particularity gives an interesting property to the states:

*Property 1.* If $s_1 \rightarrow ... \rightarrow s_n$ is a linear path, i.e., $\{s_1,..,s_n\}$ is a set of states, and $s_i$ is the state generated at the step i. If $s_1, \ldots, s_n$ are all distinct($s_1 \neq s_2 \neq ... \neq s_n$), then the time t of the state generation becomes an unique identifier of a state.

The principle of the LTL translation in ASP, is to use, at the maximum, this special case of equivalence between state and generation time. In fact, from a given initial state, we generate a path until we find a loop, or until the time bound is reached in bounded model-checking (proof in annexe). We use the wanted LTL properties as constraints on the path generation, so that each answer set is a possible combination of interactions that validates the LTL properties (to reduce the run time we can ask for a limited number of answer set).

*Example 1.* An ASP program which uses our translation to check the possibles complete networks of the second BN of Figure 2. Here we add a constraint which state that starting from (110): $x_3$ should not be true in the future. The ASP program will output one answer set where $x_2$ inhibits $x_3$.

```
%time limit
```

**Translation of LTL in ASP**
See Section. 2.2 for the formal definition.

```
%%Definition of the atomic formulas.
    phi(T) :- x_i(T).
    not_phi(T) :- not x_i(T).
%%phi1 and phi2 can also be two LTL sub-formulas.
    phi1_and_phi2(T) :- phi1(T) , phi2(T).
    phi1_or_phi2(T) :- phi1(T).
    phi1_or_phi2(T) :- phi2(T).
    phi1_Imply_phi2(T) :- not phi1_and_not_phi2(T).
    Xphi(T) :- phi(T+1) , t(T+1).
%%With tmax the bound of the steps.
    Xphi(T) :- T==Tmax, loop(T_,Tmax), Xphi(T_), t(T_).
    phi1_U_phi2(T) :- phi2(T).
    phi1_U_phi2(T) :- phi1(T) , phi1_U_phi2(T+1) , t(T+1).
    phi1_U_phi2(T) :- t(T), T==Tmax, t(T_), loop(T_,Tmax), phi1_U_phi2(T_).
%%The other formulas are given by:\\
    Gphi(T) :- not Fnotphi(T).
    Fphi(T) :- True_U_phi(T).\\
    phi1_R_phi2(T) :- not  not_varphi1_U_not_phi2(T)
```

```
t(0..8).
%initial state: variable(0|1,t), here (110)
x1(1,0).
x2(1,0).
x3(0,0).
%Incertitude on the interaction of x2 on x3
x2activateX3 :- not x2inhibateX3, not x2noeffectonX3.
x2inhibateX3 :- not x2activateX3, not x2noeffectonX3.
x2noeffectonX3 :- not x2inhibateX3, not x2activateX3.
% Transitions rules of the Boolean network
x3(1,T+1) :- x1(1,T), x2(1,T), x2activateX3, t(T).
x3(1,T+1) :- x1(1,T), x2(0,T), x2inhibateX3, t(T).
x3(1,T+1) :- x1(1,T), x2noeffectonX3, t(T).
x1(0,T) :- not x1(1,T),t(T).
x2(0,T) :- not x2(1,T),t(T).
x3(0,T) :- not x3(1,T),t(T).
loop(T,T_) :- t(T), t(T_), T<T_, x1(X1,T), x2(X2,T), x3(X3,T),
              x1(X1,T_), x2(X2,T_), x3(X3,T_).
%constraints
fx3(T) :- x3(1,T).
fx3(T) :- fx3(T+1), t(T).
fx3(T) :- t(T), T==8, t(T_), loop(T_,8), fx3(T_).
:- fx3(0).
```

### 3.2 Inference and CTL model-checking in ASP

For the CTL properties, unlike in LTL, there is no identification between a state and a step. For this reason, a translation similar to the LTL cannot be provided for all the CTL formulas.
The tree computation can be divided in n different paths from the initial state to a leaf. On each path, it is possible to check fully existential properties (no sub formulas A$\varphi$) as checking LTL formulas on a path (see proof in annexe): for example F$\varphi$ becomes EF$\varphi$. In fact, during the computation of the tree we equate an answer set to a path satisfying all the fully existential properties.
In the other case, we need the transition system to check the CTL properties

(validity of the translation in annexe):
In bounded model-checking, the program generates all the paths of fixed length k from the top to the leafs of the tree. Then it reconstructs the corresponding explored part of the state-transition model. Finally, it verifies if the model satisfies or not the CTL formulas. As we can see to find a complete model, there is one execution of the programs for each possible set of interactions.
In non bounded model-checking, we generate directly the whole states-transitions model of a given Boolean network in an answer set. Then we can directly constraint the generated model with the satisfaction of the CTL formulas. The generation cost of the transition system is exponential (we need to generate at least one transition by state so $2^n$).

**Translation of CTL in ASP**
See Section. 2.2 for the formal definition.

```
%%phi is a CTL formulas.
%%phi can be a atomic formulas phi as xi(S) or not_xi(S), i in [1..n]
    phi(S) :- state(x1,..,xi=1,..,xn) = S.
    notphi(S) :- state(x1,..,xi=0,..,xn) = S.
%%phi1 and phi2 can also be two CTL sub-formulas.
    phi1_and_phi2(S) :- phi1(S), phi2(S).
    phi1_or_phi2(S) :- phi1(S).
    phi1_or_phi2(S) :- phi2(S).
not_phi(S) :- not phi(S), state(S).
    phi1_Imply_phi2(S) :- not phi1_and_notphi2(S), state(S).
%%transition(S,S') means there is a transition from S to S'.
    EXphi(S) :- phi(S'), transition(S,S').
    Ephi1_U_phi2(S) :- phi2(S).
    Ephi1_U_phi2(S) :- phi1(S), Ephi1_U_phi2(S'), transition(S,S').
%%For EG(S) we need to cut the loop inside the formulas.
%%A naive translation would be:
    EGphi(S) :- phi(S), EGphi(S'), transition(S,S').
%%However we can easily see that there is a circular dependence if there is a loop.
%%Then we add hypothesis on EGphi:
    hypTrueEGphi(S) :- phi(S), not hypFalseEGphi(S).
    hypFalseEGphi(S) :- not hypTrueEGphi(S).
    EGphi(S) :- phi(S), hypTrueEGphi(S'), transition(S,S').
%%Finally we must add constraints to eliminate the false hypothesis.
    :- hypTrueEGphi(S), not EGphi(S), state(S).
%%From those formulas any others CTL formulas can be define.
    AXphi(S) :- not EXnotphi(S).
    AGphi(S) :- not EFnotphi(S).
    AFphi(S) :- not EGnotphi(S).
    EFphi(S) :- Etrue_U_phi(S).
    Aphi1_U_phi2(S) :- not E(notphi1_U_(notphi1_and_notphi2))_and_notEGnotphi2(S).
```

## 4 Learning Boolean networks

In some case there is no existing models and we need to learn a model from experimental data, or constraints, given as temporal logic properties. In these section, we show how to construct from scratch a Boolean network satisfying a given set of LTL properties. To learn this network, we will generate a possible execution of the Boolean network consistent with the LTL properties. The techniques is similar to the inference problem, but we only focus on generating correct trajectories. A valid execution is defined as follows:

**Definition 1.** *If p is a path of length k, i.e., a set of ordered states* $s_1 \rightarrow s_2 \rightarrow \ldots \rightarrow s_k$*. p is a valid execution of a synchronous deterministic Boolean network, if all the states* $(s_1, \ldots, s_{k-1})$ *are distinct. This means there is no loop between the step* 1 *and* $k-1$*, because the execution is synchronous and deterministic.*

*Example 2.* If the length of the paths is $k = 4$, then the following paths are valid execution: (000)→(010)→(011)→(110) or (000)→(010)→(011)→(011)
However, the following path is not a valid execution because of the (010)→(010) loop: (000)→ (**010**) → (**010**) →(110).

Here the problem is to find all valid executions satisfying the whole set of constraints. However, some set of constraints can be inconsistent. The inconsistency can be distinct in two case: two constraints can be inconsistent on a common path, we will call them path-inconsistent (resp. path-consistent), like in example 3. Or they can be inconsistent on the whole state-transition graph, like: *the next state of (100) is (000)* and *the next state of (100) is (111).* To determine this path-consistency, we can check all the possible combinations of the constraints until we find one with a satisfying model. In theory, we would generate all the possible combinations of the sets of constraints. In practice, we consider those sets of constraints given as input, because one set is the result of one experience.

*Example 3.* In the following example, we will learn a Boolean network with 3 nodes(p, q, r) from the following LTL properties:
`FGqr(000)`: in the future of (000) there is a state s, with `Gqr(s)`. `Gqr(s)` means that all states in the future of s will verify q is true and r is true (so the states (011) or (111)).
`X(110)(110)`: (110) is the immediate next state of (110).
`X(101)(001)`: (101) is the immediate next state of (001).
`F(111)(001)`: (111) is reachable in the future of (001).
We can see that `FGqr(000)` and `X(110)(110)` are not path- consistent, because `X(110)(110)` imply a loop on (110) and `FGqr` imply a loop with (111) and/or (011): this is not possible in one valid execution, we can divide the constraint in two set.
Arbitrary we use the following division: {`FGqr(000)`, `X(101)(001)`, `F(111)(001)`} on one part, and {`X(110)(110)`} on the other part.
Then the program gives a minimal valid execution for the first set of constraints: (001)→(101)→(000)→(111)→(111).
And for the other set: (110)→(110).
The transitions described by these two valid executions are consistent. But they do not represent all the possible transitions of the system: only the ones useful to satisfy the two sets of constraints. To retrieve the rules of the Boolean function, the valid executions can given as input to the algorithm described in [12]. This algorithm will compute the rules defining the Boolean network and will complete the Boolean function by putting default transition to (000) for the states without explicit transition. The state transitions corresponding to the learned Boolean network can be seen in Figure 3.

*Example 4.* An ASP program which uses our translation to solve the problem of Example 3. Here we search for a path of length at most 4. This program will output multiple answer sets, and one of them corresponds to the state-transitions graph of Figure 3.

```
% Path length
t(0..4).
% State generation
p(1,T) :- not p(0,T), t(T).
p(0,T) :- not p(1,T), t(T).
q(1,T) :- not q(0,T), t(T).
q(0,T) :- not q(1,T), t(T).
r(1,T) :- not r(0,T), t(T).
r(0,T) :- not r(1,T), t(T).
% Properties
xp_qr(T) :- r(1,T+1), p(1,T+1), q(0,T+1), t(T+1),t(T).
qr(T) :- q(1,T), r(1,T).
notqr(T) :- not qr(T), t(T).
fnotqr(T) :- notqr(T).
fnotqr(T) :- fnotqr(T+1), t(T).
gqr(T) :- not fnotqr(T), t(T).
fGqr(T) :- gqr(T).
fGqr(T) :- fGqr(T+1), t(T).
fpqr(T) :- p(1,T), q(1,T), r(1,T).
fpqr(T) :- fpqr(T+1), t(T).
% Constraints
property1 :- fGqr(T), p(0,T), q(0,T), r(0,T).
property2 :- fpqr(T), p(0,T), q(0,T), r(1,T).
property3 :- xp_qr(T), p(0,T), q(0,T), r(1,T).
%Final loop
property4 :- p(X,T), q(Y,T), r(Z,T),
p(X,4), q(Y,4), r(Z,4),
t(T), T<4.
%Inside loop
fail :- p(X,T), q(Y,T), r(Z,T),
p(X,T_), q(Y,T_), r(Z,T_),
t(T), t(T_), T<T_, T_<4.
succeed :- property1, property2, property3, property4.
fail :- not succeed.
:- fail.
```

## 5 Experiments

In this section we evaluate the performance of our method on some Boolean Networks benchmarks from the bioinformatics literature. These benchmarks are Boolean networks taken from Dubrova and Teslenko [7], which include those networks for control of over morphogenesis in Arabidopsis thaliana, budding yeast cell cycle regulation, fission yeast cell cycle regulation and mammalian cell cycle regulation. The experiments were done on a processor intel core i7 720QM 1.6GHZ, with 4Gb of RAM. The ASP solver used in these experiments is clingo 3.05 win64 [8].

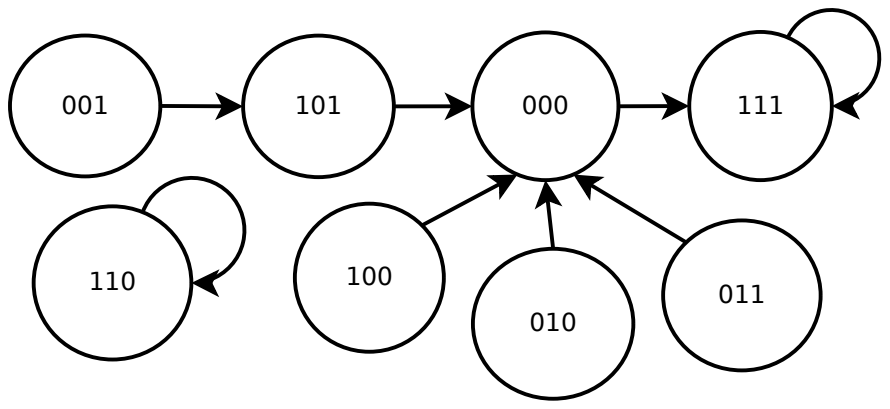

**Fig. 3.** State-transition graph of the Example 3. The two generated valid execution are presents, the others states have the default transition to (000).

The main cost of the inferring method is the model-checking of the possible networks. The tested networks are complete biological Boolean networks where some interactions are made ambiguous. We used 6 models of 10, 15, 19[2], 23, and 40 nodes. The experiments have been divided in two parts: the model-checking of asynchronous and synchronous version of the Boolean networks.

**Asynchronous Boolean network**

Table 1 shows the run time for the creation of the full transition system of a Boolean network and its non bounded model-checking. The runtime (grounding + solving) is exponential: according to the definition of asynchrony there is $n2^n$ possible state transitions for a complete asynchronous BN of $n$ nodes. However, once this generation has been done, we can easily check any other CTL formulas.

| nodes | 10 | 15 | 19 | 40 |
|---|---|---|---|---|
| Runtime(s) | 1.1s | 181s | out of memory | out of memory |

**Table 1.** Not bounded model-checking of Asynchronous Boolean network

For the bounded model-checking (see Fig. 4) we evaluate the maximum depth of exploration in one minute. Bounded model-checking is often used for searching counter example, and in our case for validating existential constraints. Those reachability problems mainly depend of the maximum depth of the search.

**Synchronous Boolean network**

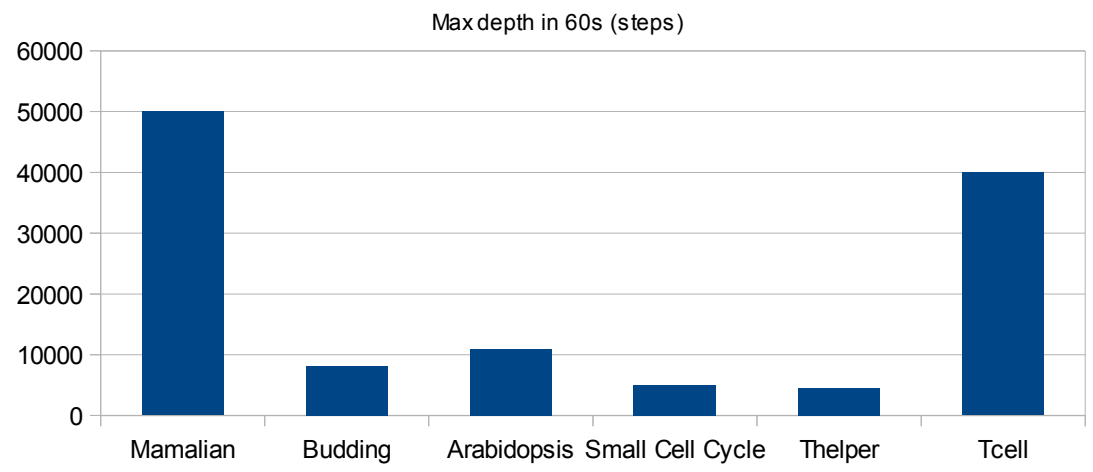

**Fig. 4.** Model-checking of synchronous Boolean network

[2] The 19 nodes model is the test example small_cell_cycle used by BIOCHAM

In the synchronous case, non bounded and bounded model-checking differ only by the loop detection. We place ourself in the worst case, and we will check the loop of size 1(unary attractor). We can see that the max depth search (see Fig 5) decreases with the number of nodes in most of the cases. However, for the 40 nodes Boolean network, the depth is far greater. In this case the rules describing the network are very simple (often one atom in the body), and the computation greatly depends on it.

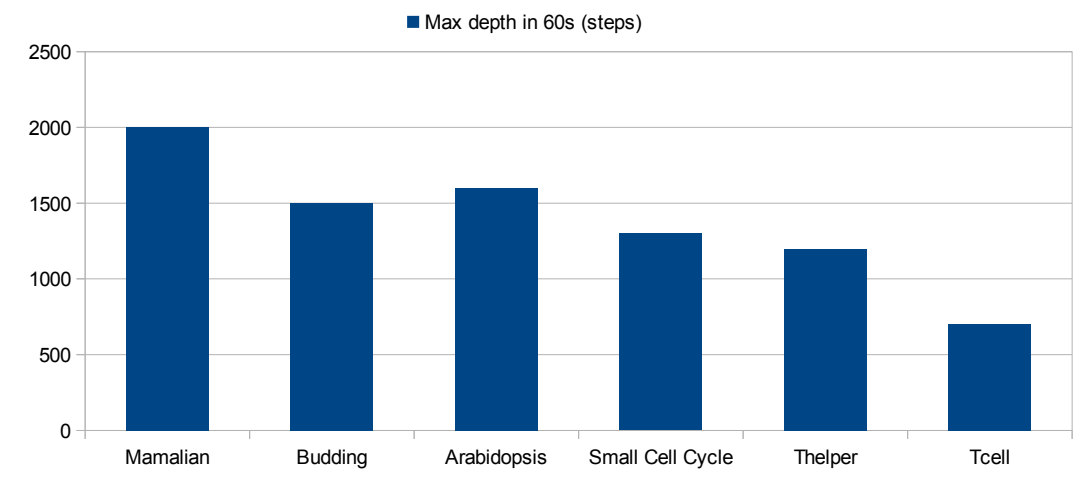

**Fig. 5.** Bounded model-checking of Asynchronous Boolean network

Finally, the computation of the Boolean networks can be optimised to use the full potential of the ASP. Moreover, the latest ASP syntax used in gringo 4.01 has shown far greater performance on the transition generation of the 15 nodes asynchronous network (75% less memory consumption, and runtime divided by 9). An optimised version of this work for new ASP syntax should approach the classical model checker performances, with greater expression power and adaptability.

## 6 Conclusion

In this work we developed techniques to infer and learn, in ASP, a Boolean network from temporal logic constraints. We gave a translation of CTL and LTL in ASP for Boolean network. These translations allow to use the expression power of the temporal logic to complete a Boolean network. The translations also allow to analyse a Boolean network with model-checking techniques. If the current learning and inferring experiments do not compete with the SAT/BDD softwares, results suggest that an ASP implementation using the new ASP definition, can reach the similar performance than the current software, with a greater adaptability and expression power. Moreover, the easy description of BN analysis, and the adaptable implementation of other methods, or temporal logics, confirms the ASP as a good programming language for the analysis of the fast evolving domain of biological models.

## References

1. Angelis, E. D., Pettorossi, A., and Proietti, M. Synthesizing concurrent programs using answer set programming. *Fundam. Inform. 120*, 3-4 (2012), 205–229.
2. Arellano, G., Argil, J., Azpeitia, E., Benítez, M., Carrillo, M. A., Góngora, P. A., Rosenblueth, D. A., Alvarez-Buylla, E. R., et al. " antelope": a hybrid-logic model checker for branching-time boolean grn analysis. *BMC bioinformatics 12*, 1 (2011), 490.
3. Calzone, L., Chabrier-Rivier, N., Fages, F., and Soliman, S. Machine learning biochemical networks from temporal logic properties. 68–94.
4. Charbon, E., Miliozzi, P., Carloni, L., Ferrari, A., and Sangiovanni-Vincentelli, A. Modeling digital substrate noise injection in mixed-signal ic's. *Computer-Aided Design of Integrated Circuits and Systems, IEEE Transactions on 18*, 3 (1999), 301–310.
5. Cimatti, A., Clarke, E., Giunchiglia, F., and Roveri, M. Nusmv: a new symbolic model checker. *International Journal on Software Tools for Technology Transfer 2*, 4 (2000), 410–425.
6. Clarke, E. M., and Emerson, E. A. Design and synthesis of synchronization skeletons using branching-time temporal logic. In *Logic of Programs* (1981), pp. 52–71.
7. Dubrova, E., and Teslenko, M. A sat-based algorithm for finding attractors in synchronous boolean networks. *IEEE/ACM Transactions on Computational Biology and Bioinformatics 8*, 5 (2011), 1393–1399.
8. Gebser, M., Kaufmann, B., Neumann, A., and Schaub, T. Conflict-driven answer set solving. In *IJCAI* (2007), vol. 7, pp. 386–392.
9. Gelfond, M., and Lifschitz, V. The stable model semantics for logic programming. In *ICLP/SLP* (1988), vol. 88, pp. 1070–1080.
10. Green, D., Leishman, T., and Sadedin, S. The emergence of social consensus in boolean networks. In *Artificial Life, 2007. ALIFE '07. IEEE Symposium on* (2007), pp. 402–408.
11. Heymans, S., Nieuwenborgh, D. V., and Vermeir, D. Synthesis from temporal specifications using preferred answer set programming. In *ICTCS* (2005), pp. 280–294.
12. Inoue, K., Ribeiro, T., and Sakama, C. Learning from interpretation transition. *Machine Learning* (2013), 1–29.
13. Kauffman, S. A. Metabolic stability and epigenesis in randomly constructed genetic nets. *Journal of theoretical biology 22*, 3 (1969), 437–467.
14. Klamt, S., Saez-Rodriguez, J., Lindquist, J. A., Simeoni, L., and Gilles, E. D. A methodology for the structural and functional analysis of signaling and regulatory networks. *BMC Bioinformatics 7* (2006), 56.
15. Lähdesmäki, H., Shmulevich, I., and Yli-Harja, O. On learning gene regulatory networks under the boolean network model. *Machine Learning 52*, 1-2 (2003), 147–167.
16. Queille, J.-P., and Sifakis, J. Specification and verification of concurrent systems in cesar. In *Symposium on Programming* (1982), pp. 337–351.
17. Tang, C. K. F., and Ternovska, E. Model checking abstract state machines with answer set programming. In *Abstract State Machines* (2005), pp. 397–416.

## Annexes

### Annexe A: Validity of the LTL translation in ASP

Given a n nodes ($x_1$, ... , $x_n$) Boolean network generated on $k$ steps, $k \in \mathbb{N}^*$. Then s(t)=($x_1$(t), ... , $x_n$(t)) is the state of the Boolean network at step t, and the generated path $p$, from $t = 0$ to $t = k$, can be written $p = (s(0), ..., s(k))$.

*Remark 1.* If $\varphi$ is a LTL property, then p $\models \varphi$ is equivalent to p(0) $\models \varphi$ (see Section. 2.2). By the same way, we note that for 0≤i≤k, $p(i) \models \varphi$ means there is a path $p_i$ with $p(i) = p_i(0)$ and $p_i \models \varphi$.

As explained in Section. 3.1, we compute the network until we find a loop, or until a given maximum runtime if there is no loop. The result of the computation can be divided in two possibility: there is no loop, and there is a loop with the last state.
In the first case, all the states of the path are distinct, and the property. 1 (see Section. 3.1) can be applied: the state $s(t)$ can be identified by the generation step $t$. Thanks to the remark. 1 we can write: for each state $p(i) \in p$, 0≤i≤, $p(i)$ $\models \varphi$ means $i \models \varphi$, which can be contracted in $\varphi$(i).
With those notations, a state $p(i) \models$X$\varphi$ iff $p(i+1) \models \varphi$ can be written X$\varphi$(i) iff $\varphi$(i+1). It is exactly the ASP translation X$\varphi$(t):- $\varphi$(t+1). By the same way, the operator U is correctly translated in ASP.
In the second case, we need to manage the loop and the non equivalence step/state. The loop only appears at the last step $k$. For 0≤i≤k-1 we have $s(0) \neq ... \neq s(i)$ $\neq ... \neq$s(k-1): We can apply the property 1 and the asp translation. For $s(k)$ we transmit the property since the behaviour have already been verified in the k previous steps. This transmission is given by the following rule in ASP, for example with X$\varphi$: Xphi(T) :- T==Tmax, loop(T_,Tmax), Xphi(T_),which means that Tmax $\models$X$\varphi$ if there is loop between T_ and Tmax (s(T_)=s(Tmax)) and T_$\models$X$\varphi$.

### Annexe B: Validity of the CTL translation in ASP

As seen in Section. 2.2, all the CTL formulas can be describe with only EX$\varphi$, E$\varphi_1$U$\varphi_2$,and EG$\varphi$.
If $s$ is a state, and $M$ a Kripke model of the Boolean network as defined in Section. 2.2. The prefix E means there are existential properties: there exits a path were the property will holds. The existence of a path $p$=($s$,$s_1$,...,$s_n$) is equivalent to the existence of transitions T($s_i$,$s_{i+1}$) with $0 \leq i \leq n-1$. From those observation we can directly translate the 3 properties:
$(M, s) \models$ EX$\varphi$ iff $\exists$ a path p | p(0)=s and $(M, s_1 = p(1)) \models \varphi$. If the Kripke model become implicit, this properties can be written EX$\varphi$(s) iff $\exists$ a transition T($s$,$s_1$) and $\varphi(s_1)$ which is equivalent to the ASP translation $EXphi(S) : -phi(S'), transition(S, S')$, with $S$, and $S'$ two states.

$(M, s) \models E\varphi_1 U\varphi_2$ iff $\exists$ a path $p \mid p(0) = s$ and $\exists i \geq 0 \mid (M, s_i = p(i)) \models \varphi_2$ and $\forall 0 \leq k \leq i$ $(M, s_k = p(k)) \models \varphi_1$. Again we can contract the formulas in:
E$\varphi_1$U$\varphi_2$(s) iff $\exists$ a set of transition T $=\{(s,s_1),...,(s_k,s_{k+1}),...(s_{i-1},s_i) \mid \varphi_2(s_i)$, and $\forall$ $0 \leq k \leq i, \varphi_1(s_k)$. Then obviously, we notice that E$\varphi_1$U$\varphi_2(s_i)$ is true, and recursively for $0 \leq k < i$ E$\varphi_1$U$\varphi_2(s_k)$ is true if E$\varphi_1$U$\varphi_2(s_{k+1})$. Those two observations are expressed by the CTL translation:
Ephi1_U_phi2(S) :- phi2(S).
Ephi1_U_phi2(S) :- phi1(S), Ephi1_U_phi2(S'), transition(S,S').
Finally, (M,s) $\models$ EG$\varphi$ iff $\exists$ a path $p \mid p(0) = s$ and $\forall$ $0 \leq i$ $(M, s_i = p(i)) \models \varphi$ can be reduced in:
EG$\varphi(s)$ iff $\exists$ a set of transition T $=\{(s,s_1),...,(s_k,s_{k+1}),...(s_{i-1},s_i) \mid \forall$ 0≤k≤i, $\varphi(s_k)$ is true.
In a finite path p=$(s_1,...,s_n)$, $\forall$ $s_i$ $\in$p, i<n : EG$\varphi(s_i)$ is true iff $\varphi(s_i)$ is true and $\exists$ T$(s_i,s_{i+1})$ with EG$\varphi(s_{i+1})$. Then EG$\varphi(s_n)$ is true iff $\varphi(s_n)$ is true.
In a non finite path p=$(s_1,...)$, The property EG$\varphi(s_1)$ holds iff there is a loop ,of size k, from $s_1$ to $s_k$ in the path, with $\varphi(s_i)$ true for all i$\in[1..k]$, and for this reason we cannot translate directly the property. To check this property we make the assumption that if $\varphi$ holds then EG$\varphi$ can be true. However, we are sure that if $\varphi$ is false then EG$\varphi$ will be false. This will create a two possible answer sets for each state where $\varphi$ holds. We define EG$\varphi$(s) true if $\varphi$(s) is true and there is a transition $t(s, s')$ with hypEG$\varphi$(s') true. All we need now is to add a constraint to eliminate the wrong assumption: :- hypTrueEGphi(S), not EGphi(S), state(S).
If the property is true, only remains the answer set with hypTrueEGphi(s) and EGphi(S) true **for all** states: this is the case were all the states verify $\varphi$, this is what we wanted.
If the property is false, there is only the answer set with hypfalseEGphi(s) for all states, and then no state with EGphi(S), this what we wanted.

# Molecular Interaction Automated Maps

Robert Demolombe, Luis Fariñas del Cerro, and Naji Obeid*

Université de Toulouse and CNRS, IRIT, Toulouse, France
robert.demolombe@orange.fr, luis.farinas@irit.fr, naji.obeid@irit.fr

**Abstract.** Determining the role of genes and their interference in a cell life cycle has been at the center of metabolic network researches and experiments. Logical representations of such networks aim to guide scientists in their reasoning in general and to help them find inconsistencies and contradictions in their results in particular. This paper presents a new logical model capable of describing both positive (activation) and negative (inhibition) reactions of metabolic pathways based on a fragment of first order logic. An efficient automated deduction method will also be introduced, based on a translation procedure that transform first order formulas into quantifier free formulas. Then questions can either be answered by deduction to predict reaction results or by abductive reasoning to infer reactions and protein states.

## 1 Introduction

Cells in general and human body cells in particular incorporate a large series of intracellular and extracellular signalings, notably protein activations and inhibitions, that specify how they should carry out their functions. Networks formed by such biochemical reactions, often referred as *pathways*, are at the center of a cell's existence and they range from simple and chain reactions and counter reactions to simple and multiple regulations and auto regulations, that can be formed by actions defined in Figure 1. Cancer, for example, can appear as a result of a pathology in the cell's pathway, thus, the study of signalization events appears to be an important factor in biological, pharmaceutical and medical researches [14, 11, 7]. However, the complexity of the imbrication of such processes makes the use of a physical model as a representation seem complicated.

In the last couple of decades, scientists that used artificial intelligence to model cell pathways [10, 9, 16, 17, 6, 21, 15] faced many problems especially because information about biological networks contained in knowledge bases is generally incomplete and sometimes uncertain and contradictory. To deal with such issues, abduction [3] as theory completion [12, 16] is used to revise the state of existing nodes and add new

* LNCSR Scholar

nodes and arcs to express new observations. Languages that were used to model such networks had usually limited expressivity, were specific to special pathways or were limited to general basic functionalities. We, in this work, present a fragment of first order logic [19] capable of representing node states and actions in term of positive and negative relation between said nodes. Then an efficient proof theory for these fragments is proposed. This method can be extended to define an abduction procedure which has been implemented in SOLAR [13], an automated deduction system for consequence finding. A previous version of this work has been presented in BIOCOMP'13 [5].
For queries about the graph that contains negative actions, it is assumed that we have a complete representation of the graph. The consequence is that the negation is evaluated according to its definition in classical logic instead of some non-monotonic logic. This approach guarantees a clear meaning of answers. Since the completion of the graph is formalized a la Reiter we used the equality predicate. It is well known that equality leads to very expensive automated deductions. This problem has been resolved by replacing completed predicates by their extensions where these predicates are used to restrict the domain of quantified variables. The result of this translation is formulated without variables where consequences can be derived as in propositional logic. This is one of the main contributions of this paper.

The rest of this paper is organized as follows. Section 2 presents a basic language and axiomatic capable of describing general pathways, ans shows how it is possible to extends this language and axiomatic to address specific and real life examples. Section 3 defines a translation procedure capable of eliminating first order variables and equality predicates and shows how it can be applied to derive new axiomatic that can be used in the automated deduction process in SOLAR. Section 4 provide some case studies, and finally section 5 gives a summary and discusses future works.

## 2 Logical Model

In this section we will present a basic language capable of modeling some basic positive and negative interaction between two or more proteins in some pathway. We will first focus on the stimulation and inhibition actions, points $(g)$ and $(i)$ of Figure 1, and then show how this language can be modified to express the different other actions described in the same figure.

### 2.1 Formal Language

Let's consider a fragment of first order logic with some basic predicates, boolean connectives $(\wedge)$ and, $(\vee)$ or, $(\neg)$ negation, $(\rightarrow)$ implication, $(\leftrightarrow)$ equivalence, $(\exists)$ existential and $(\forall)$ universal quantifiers, and $(=)$ equality.

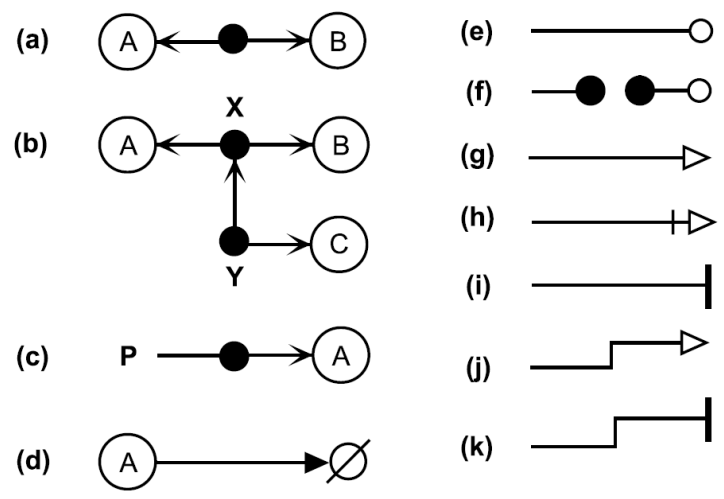

Fig. 1: Symbol definitions and map conventions.

($a$) Proteins A and B can bind to each other. The node placed on the line represents the A:B complex. ($b$) Multimolecular complexes: $x$ is A:B and $y$ is(A:B):C. ($c$) Covalent modification of protein A. ($d$) Degradation of protein A. ($e$) Enzymatic stimulation of a reaction. ($f$) Enzymatic stimulation in transcription. ($g$) General symbol for stimulation. ($h$) A bar behind the arrowhead signifies necessity. ($i$) General symbol for inhibition. ($j$) Shorthand symbol for transcriptional activation. ($k$) Shorthand symbol for transcriptional inhibition.

The basic state predicates are $A(x)$, $I(x)$ and $P(x)$ respectively meaning that the protein $x$ is *Active*, *Inhibited* or *Present*. And the basic state axioms that indicate that a certain protein $x$ can never be in both *Active* and *Inhibited* states at the same time are:

$$A(x) \rightarrow \neg I(x) \ . \qquad (1) \qquad\qquad \neg A(x) \rightarrow I(x) \vee \neg P(x) \ . \qquad (2)$$

$$I(x) \rightarrow \neg A(x) \ . \qquad (3) \qquad\qquad \neg I(x) \rightarrow A(x) \vee \neg P(x) \ . \qquad (4)$$

An interaction between two or more different proteins is expressed by a predicate of the form $Action(protein_1, ..., protein_n)$. In our case we are interested by the simple *Activation* and *Inhibition* actions that are defined by the following predicates:

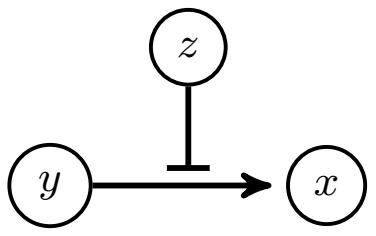

Fig. 2: Activation

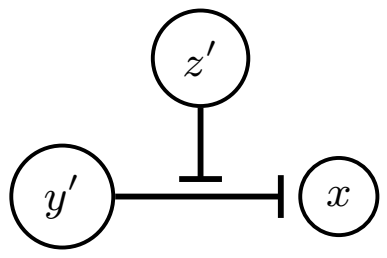

Fig. 3: Inhibition

- $CAP(y, x)$: the *Capacity of Activation* expresses that the protein $y$ has the capacity to activate the protein $x$.
- $CICAP(z, y, x)$: the *Capacity to Inhibit the Capacity of Activation* expresses that the protein $z$ has the capacity to inhibit the capacity of the activation of $x$ by $y$.
- $CIP(y', x)$: the *Capacity to Inhibit a Protein* expresses that the protein $y'$ has the capacity to inhibit the protein $x$.

– $CICIP(z', y', x)$: the *Capacity to Inhibit the Capacity of Inhibition of a Protein* expresses that the protein $z'$ has the capacity to inhibit the capacity of inhibition of $x$ by $y'$.

In the next section we will define the needed axioms that will be used to model the *Activation* and *Inhibition* actions.

### 2.2 Action axioms

Given the fact that a node can acquire the state active or inhibited depending on different followed pathways, one of the issues answered by abduction is to know which set of proteins is required to be active of inhibited for our target protein be active or inhibited.

**Axiomatic of activation:** A protein $x$ is active if there exists at least one *active* protein $y$ that has the capacity to activate $x$, $CAP(y, x)$, and for every protein $z$ that has the capacity to inhibit the capacity of activation of $x$ by $y$, $CICAP(z, y, x)$, $z$ is *not active*. (Figure 2)

$$\forall x(\exists y(A(y) \land CAP(y, x) \land \forall z(CICAP(z, y, x) \rightarrow \neg A(z))) \rightarrow A(x)) \ . \tag{5}$$

**Axiomatic of inhibition** A protein $x$ is inhibited if there exists at least one *active* protein $y'$ that has the capacity to inhibit $x$, $CIP(y', x)$, and for every protein $z'$ that has the capacity to inhibit the capacity of inhibition of $x$ by $y'$, $CICIP(z', y', x)$, $z'$ is *not active*. (Figure 3)

$$\forall x(\exists y'(A(y') \land CIP(y', x) \land \forall z'(CICIP(z', y', x) \rightarrow \neg A(z'))) \rightarrow I(x)) \ . \tag{6}$$

### 2.3 Extension with new states and actions

The basic language defined in 2.1 and 2.2 can be easily extended to express different and more precise node statuses and actions. For example the action of *phosphorylation* can be defined by the following predicates:

– $CP(z, y, x)$: the *Capacity of Phosphorylation* expresses that the protein $z$ has the capacity to phosphorylate the protein $y$ on a certain site, knowing that $x$ is the result of said phosphorylation.
– $CICP(t, z, y, x)$: the *Capacity to Inhibit the Capacity of Phosphorylation* expresses that the protein $t$ has the capacity to inhibit the capacity of the phosphorylation of $y$ by $z$ leading to $x$.

We can now define the new phosphorylation axiom as:

$$\forall x(\exists y_1 \exists y_2(A(y_1) \land A(y_2) \land CP(y_1, y_2, x) \land \forall z(CICP(z, y_1, y_2, x) \rightarrow \neg A(z))) \rightarrow A(x)) \ .$$

## 3 Automated Deduction Method

In this section we define a fragment of first order logic with constants and equality, and without functions, that is a special case of *Evaluable* formulas [2] and *Domain Independent* formulas [22], and a generalization of *Guarded* formulas [1] called *Restricted* formulas. The properties of this fragment allow us to define a procedure capable of eliminating the quantifiers in this fragment, in other words to transform the first order formulas in formulas without variables, in order to obtain an efficient automated deduction procedure with these fragments.

**Definition 1.** *$Domain formulas$ are defined by the following grammar:*

$$\delta ::= P(\overline{x}, \overline{c}) | \varphi \lor \psi | \varphi \land \psi | \varphi \land \neg\psi \ . \tag{7}$$

Where variables $\overline{x}$ and constants $\overline{c}$ denote $x_1, ..., x_n$ and $c_1, ..., c_m$ respectively. The set of free variables in $\varphi$ is the same as the set of free variables in $\psi$ for $\varphi \lor \psi$, and the set of free variables in $\psi$ is included in the set of free variables in $\varphi$ for $\varphi \land \neg\psi$[1].

**Definition 2.** *Restricted formulas are formulas without free variables defined by the following grammar:*

$$\delta ::= \forall\overline{x}(\varphi \to \psi) | \exists\overline{x}(\varphi \land \psi) \ . \tag{8}$$

Where $\varphi$ is a domain formula and $\psi$ is either a restricted formula or a formula without quantifiers, and every variable appearing in a restricted formula must appear in a domain formula. The set of variables in $\overline{x}$ is included in the set of free variables in $\varphi$; The same goes for $\psi$.
Examples: $\forall x(P(x) \to Q(x))$. $\forall x(P(x) \to \exists y(Q(y) \land R(x, y)))$.

**Definition 3.** *A completion formula is a formula of the following form:*

$$\begin{aligned}\forall x_1, ..., x_n \ (P(x_1, ..., x_n, c_1, ..., c_p) \leftrightarrow ((x_1 = a_{1_1} \land ... \land x_n = a_{1_n}) \lor ... \lor \\ (x_1 = a_{m_1} \land ... \land x_n = a_{m_n}))) \ .\end{aligned} \tag{9}$$

Where $P$ is a predicate symbol of arity $n + p$, and $a_i$ are constants. Completion formulas are similar to the completion axioms defined by Reiter in [18] where the implication is substituted by an equivalence.

**Definition 4.** *Given a domain formula $\varphi$ and a set of completion formulas $\alpha_1, ..., \alpha_n$ such that for each predicate symbol in $\varphi$ there exists a completion formula $\alpha$ for this*

[1] There are no special constraints for $\varphi \land \psi$.

*predicate symbol, we say that the set of completion formulas $\alpha_1, ..., \alpha_n$ covers $\varphi$ and will be noted $C(\varphi)$*[2]*.*

**Definition 5.** *Given a domain formula $\varphi$, we define the domain of the variables of $\varphi$, denoted $D(\mathcal{V}(\varphi), C(\varphi))$, as follows:*

– if $\varphi$ is of the form $P(x_1, ..., x_n, c_1, ..., c_p)$, and $C(\varphi)$ of the form:

$$\forall x_1, ..., x_m (P(x_1, ..., x_m, c_1, ..., c_l) \leftrightarrow ((x_1 = a_{1_1} \wedge ... \wedge x_m = a_{1_m}) \vee ... \vee (x_1 = a_{q_1} \wedge ... \wedge x_m = a_{q_m}))) \ .$$

where $n \leq m$ and $l \leq p$.

$$\text{then } D(\mathcal{V}(\varphi), C(\varphi)) = \{< a_{1_1}, ..., a_{1_n} >, ..., < a_{q_1}, ..., a_{q_n} >\} \ . \quad (10)$$

– if $\varphi$ is of the form $\varphi_1 \vee \varphi_2$ then:

$$D(\mathcal{V}(\varphi_1 \vee \varphi_2), C(\varphi_1 \vee \varphi_2)) = D(\mathcal{V}(\varphi_1), C(\varphi_1)) \cup D(\mathcal{V}(\varphi_2), C(\varphi_2)) \ . \quad (11)$$

– if $\varphi$ is of the form $\varphi_1 \wedge \varphi_2$ then:

$$D(\mathcal{V}(\varphi_1 \wedge \varphi_2), C(\varphi_1 \wedge \varphi_2)) = D(\mathcal{V}(\varphi_1), C(\varphi_1)) \otimes_c D(\mathcal{V}(\varphi_2), C(\varphi_2)) \ . \quad (12)$$

Where $\otimes_c$ [22] is a join operator and $c$ is a conjunction of equalities of the form $i = j$ the same variable symbol appears in $\varphi_1 \wedge \varphi_2$ in position $i$ in $\varphi_1$ and in position $j$ in $\varphi_2$.

– if $\varphi$ is of the form $\varphi_1 \wedge \neg\varphi_2$ then:

$$D(\mathcal{V}(\varphi_1 \wedge \neg\varphi_2), C(\varphi_1 \wedge \neg\varphi_2)) = D(\mathcal{V}(\varphi_1), C(\varphi_1)) \setminus D(\mathcal{V}(\varphi_1 \wedge \varphi_2), C(\varphi_1 \wedge \varphi_2)). \quad (13)$$

Where $\setminus$ denotes the complement of the domain of each shared variable of $\varphi_2$ with respect to $\varphi_1$.

*Example 1.* Considering the three domains formulas $P(x)$, $Q(x)$, $R(x, y)$ and their corresponding completion formulas as following:

$\forall x(P(x) \rightarrow x = a \vee x = d)$ we have $D(\mathcal{V}(P(x)), C(P(x))) = \{< a >, < d >\}$ .

$\forall x(Q(x) \rightarrow x = b \vee x = c)$ we have $D(\mathcal{V}(Q(x)), C(Q(x))) = \{< b >, < c >\}$ .

$\forall x, y(R(x, y) \rightarrow (x = a \wedge y = b) \vee (x = a \wedge y = c) \vee (x = b \wedge y = e))$ we have

$D(\mathcal{V}(R(x, y)), C(R(x, y))) = \{< a, b >, < a, c >, < b, e >\}$ .

[2] It may be that for some predicate, or some atomic formula, there is no completion formula. In that case $C(\varphi)$ is not defined. For instance, if for the predicate $P$ we only have $\alpha$ : $\forall y(P(c_2, y) \leftrightarrow y = c_3)$, there is no completion formula for $P(x_1, c_1)$ while there is a completion formula for $P(c_2, x_2)$.

If we have:

$\varphi_1 = P(x) \vee Q(x)$ then $D(\mathcal{V}(\varphi_1), C(\varphi_1)) = \{< a >, < b >, < c >, < d >\}$ .

$\varphi_2 = R(x, y) \wedge P(x)$ then $D(\mathcal{V}(\varphi_2), C(\varphi_2)) = \{< a, b >, < a, c >\}$ .

$\varphi_3 = R(x, y) \wedge \neg P(x)$ then $D(\mathcal{V}(\varphi_3), C(\varphi_3)) = \{< b, e >\}$ .

**Quantifier elimination procedure**

Let $\varphi$ be a restricted formula of the following forms: $\forall\overline{x}(\varphi_1(\overline{x}) \rightarrow \varphi_2(\overline{x}))$ or $\exists\overline{x}(\varphi_1(\overline{x}) \wedge \varphi_2(\overline{x}))$, let $C(\varphi_1(\overline{x}))$ a set of completion formulas for $\varphi_1$, then we define recursively a translation $T(\varphi, C(\varphi))$, allowing to replace universal (existential) quantifiers by conjunction (disjunction) of formulas where quantified variables are substituted by constants as follows:

– if $D(\mathcal{V}(\varphi_1), C(\varphi_1)) = \{< \overline{c_1} >, ..., < \overline{c_n} >\}$ with $n > 0$:

$$T(\forall\overline{x}(\varphi_1(\overline{x}) \rightarrow \varphi_2(\overline{x})), C(\varphi)) = T(\varphi_2(\overline{c_1}), C(\varphi_2(\overline{c_1}))) \wedge ... \wedge T(\varphi_2(\overline{c_n}), C(\varphi_2(\overline{c_n}))) \ .$$

$$T(\exists\overline{x}(\varphi_1(\overline{x}) \wedge \varphi_2(\overline{x})), C(\varphi)) = T(\varphi_2(\overline{c_1}), C(\varphi_2(\overline{c_1}))) \vee ... \vee T(\varphi_2(\overline{c_n}), C(\varphi_2(\overline{c_n}))) \ .$$

– if $D(\mathcal{V}(\varphi_1), C(\varphi_1)) = \varnothing$ :

$$T(\forall\overline{x}\ (\varphi_1(\overline{x}) \rightarrow \varphi_2(\overline{x}))\ ,\ C(\varphi)) = True. \qquad T(\exists\overline{x}\ (\varphi_1(\overline{x}) \wedge \varphi_2(\overline{x}))\ ,\ C(\varphi)) = False.$$

*Note 1.* It is worth nothing that in this translation process each quantified formula is replaced in the sub formulas by constants. The consequence is that if a sub formula of a restricted formula is of the form $\forall\overline{x}(\varphi_1(\overline{x}) \rightarrow \varphi_2(\overline{x}, \overline{y}))$ or $\exists\overline{x}(\varphi_1(\overline{x}) \wedge \varphi_2(\overline{x}, \overline{y}))$ where the quantifiers $\forall\overline{x}$ or $\exists\overline{x}$ are substituted by their domain values, the variables in $\overline{y}$ must have been already substituted by its corresponding constants.

Then in the theory $\mathcal{T}$ in which we have the axioms of equality and axioms of the form $\neg(a = b)$ for each constant $a$ and $b$ representing different objects, which are called unique name axioms by Reiter in [18], we have the following main theorem:

**Theorem 1.** *Let $\varphi$ be a restricted formula, and $C(\varphi)$ a completion set of formulas of the domain formulas of $\varphi$, then:*

$$\mathcal{T},\ C(\varphi) \vdash \varphi \leftrightarrow T(\varphi, C(\varphi)) \ . \tag{14}$$

*Proof.* The proof consists of applying induction on the number of domain formulas in a restricted formula to prove that the theorem holds for any number domain formulas.

*Example 2.*

Let's consider the case where a protein $b$ has the capacity to activate another protein $a$, and that two other proteins $c_1$ and $c_2$ have the capacity to inhibit the capacity of activation of $a$ by $b$. This proposition can be expressed by the following completion axioms:

– $\forall y(CAP(y,a) \leftrightarrow y = b)$: Where $b$ is the only protein that has the capacity to activate $a$.
– $\forall z(CICAP(z,b,a) \leftrightarrow z = c_1 \vee z = c_2)$: Where $c_1$ and $c_2$ are the only proteins that have the capacity to inhibit the capacity of activation of $a$ by $b$.

Using the activation axiom defined in section 2 and the translation procedure, we can deduce $A(b) \wedge \neg A(c_1) \wedge \neg A(c_2) \wedge \rightarrow A(a)$. Which means that the protein $a$ is active if the protein $b$ is active and the proteins $c_1$, $c_2$ are not active.

Let's also consider that a protein $d$ has the capacity to inhibit the protein $a$ and that there is no proteins capable of inhibiting the capacity of inhibition of $a$ by $d$. This proposition can be expressed by the following completion axioms:

– $\forall y(CIP(y,a) \leftrightarrow y = d)$: Where $d$ is the only protein that has the capacity to inhibit $a$.
– $\forall z(CICIP(z,d,a) \leftrightarrow false)$: Where there are no proteins capable of inhibiting the capacity of inhibition of $a$ by $d$.

Using the previous inhibition axiom and these completion axioms we can deduce $A(d) \rightarrow I(a)$. Which means that the protein $a$ is inhibited if the protein $d$ is active.

## 4 Automated Reasoning

From what we defined in sections 2 and 3, the resulting translated axioms are of the following type *conditions* $\rightarrow$ *results*, and can be chained together to create a series of reactions forming our pathway. Then questions of two different types can be answered using *abductive* or *deductive* reasoning.

(a) Questions answered by *abduction* looks for minimal assumptions ($H$) that must be added to the knowledge base ($T$) to derive that a certain fact ($C$) is true. A question can be of the following form: *what are the proteins and their respective states (active or inhibited) that should be present in order to derive that a certain protein is active or inhibited.*
(b) Questions answered from an abduced hypothesis $H$, that we call *test basis* and will be noted $TB_H$, are minimal consequences that can be derived by *deduction* over $T$ and $H$, knowing that they are not direct consequences of $T$ nor they can be deduced by $H$.

Both $H$ and $TB_H$ types of questions can be addressed in SOLAR (SOL for Advanced Reasoning) [13] a first-order clausal consequence finding system based on SOL (Skip Ordered Linear) tableau calculus [8, 20]. We will now present two examples corresponding to those two types of questions.

*Example 3.*

In the following we are going to show an example, based on figure 4, demonstrating abduction type queries where three coherent pathways have been found [11]. From section 2.3 we can define new predicates to suit the needs of the pathway, as the *Capacity of Binding* $CB(z, y, x)$ and the *Capacity to Inhibit the Capacity of Binding* $CICB(t, z, y, x)$. These new predicates can be used to model the binding between p53 and Bak using the predicate $CP(p53, bak, p53_bak)$ where p53_bak is the complex formed by such binding.

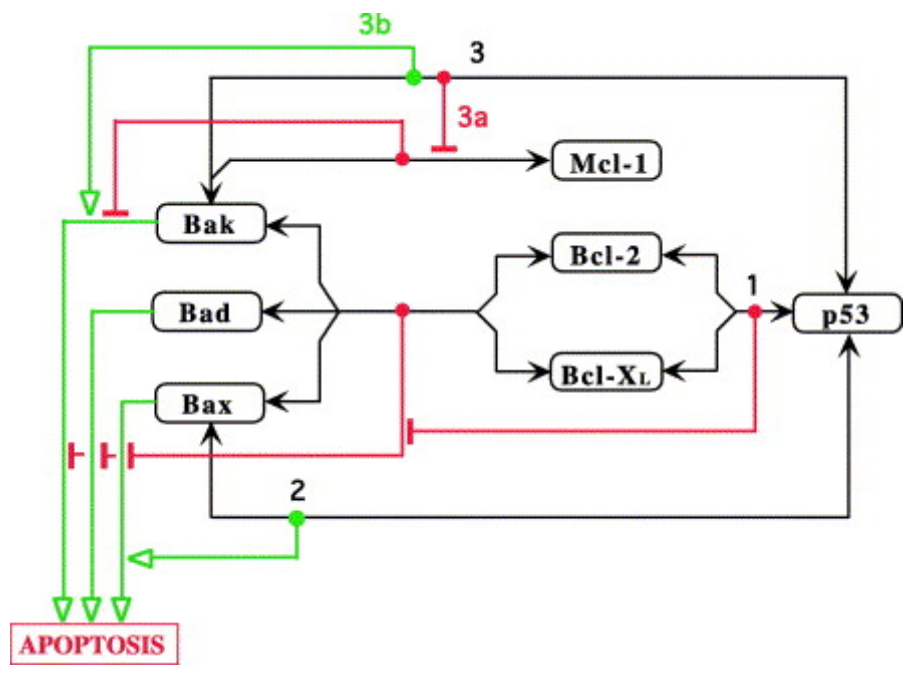

Fig. 4: Mitochondrial apoptosis induced by p53 independently of transcription

With these new predicates, new axioms can be defined that would enrich the descriptive capacities of the old ones, as seen in 2.3. Then the translation procedure applied to these axioms and to the completion axioms can be of the following form:

1. $A(p53) \wedge A(bak) \rightarrow A(bak_p53)$. Where $bak_p53$ is the result of the binding between p53 and Bak.
2. $A(bak_p53) \rightarrow I(bak_mcl)$. Where $bak_mcl$ is the result of binding between Bak and Mcl-1.
3. $A(bak_p53) \wedge \neg A(b_complex) \wedge \neg A(bak_mcl) \rightarrow A(apoptosis)$. Where $b_complex$ is result of the binding between Bcl-2, Bcl-XL, Bak, Bad, and Bax.
4. $A(bak) \wedge \neg A(b_complex) \wedge \neg A(bak_mcl) \rightarrow A(apoptosis)$
5. $A(p53) \wedge A(bcl) \rightarrow A(p53_bb_complex)$. Where $bcl$ represents Bcl-2 and Bcl-XL. $p53_bb_complex$ is the result of binding between p53, Bcl-2 and Bcl-XL.

6. $A(p53_bb_complex) \rightarrow I(b_complex)$
7. $A(bax) \wedge \neg A(b_complex) \rightarrow A(apoptosis)$
8. $A(p53) \wedge A(bax) \wedge \neg A(b_complex) \rightarrow A(apoptosis)$
9. $A(bad) \wedge \neg A(b_complex) \rightarrow A(apoptosis)$

If we want to know what are the proteins and their respective states that should be present in order to derive that the cell reached apoptosis, the answer is given by applying abduction over the previous set of compiled clauses. In the set of consequences returned by SOLAR we can find the following:

– $A(p53) \wedge A(bcl) \wedge A(bak)$: is a plausible answer, because p53 can bind to Bcl giving the *p53_bb_complex*, which can in return inhibit the *b_complex* that is responsible of inhibiting the capacity of Bak to activate the cell's apoptosis.
– Another interpretation of the previous answer is that p53 can also bind to Bak giving the *bak_p53* protein, which can in return inhibit the *bak_mcl* responsible of inhibiting the capacity of Bak to activate the cell's apoptosis. *bak_p53* can also stimulate Bak to reach apoptosis. Without forgetting that *p53_bb_complex* inhibit *b_complex*.

*Example 4.*

Let's consider the case where proteins $b$ and $c$ have the capacity to activate the protein $a$, $b$ can also inhibit $d$, and $e$ can inhibit $b$. This proposition can be expressed by the following completion axioms $T$:

$$A(b) \rightarrow A(a). \quad A(c) \rightarrow A(a). \quad A(b) \rightarrow I(d). \quad A(e) \rightarrow I(b).$$

In order to derive $A(a)$, one the following hypotheses $H$ should be considered: $A(b)$ or $A(c)$. For $H = A(b)$ we can deduce the following $TB_{A(b)}$ consistency conditions: $\neg A(e)$ and $\neg A(d)$, that describe that for $A(b)$ to be consistent with the main proposition, which is $A(a)$, as a condition the protein $e$ should not be active, and as a result the protein $d$ is inhibited (not active). These new conditions can help us reason about consistency because if we know, by means of scientific experiments or as a result of some observations, that either $d$ or $e$ is active, this means that $b$ is not active, which leaves us with $c$ as the only protein that activates $a$.

## 5 Conclusion

A new language has been defined in this paper capable of modeling both positive and negative causal effects between proteins in a metabolic pathway. We showed how this basic language can be extended to include more specific actions that describes different relations between proteins. These extensions are important in this context, because there is always the possibility that new types of actions are discovered through biological experiments. We later showed how the axioms defined in such languages can be compiled against background knowledge, in order to form a new quantifier free axioms that could be used in either deduction or abduction reasoning. Although the first order axioms can be also well used to answer queries by deduction or abduction methods, the main advantage of translated axioms is their low computation time needed in order to derive consequences.

Future works can focus on extending the language used to define domain formulas, introducing the notion of time and quantities in the model. Trying to get as precise as possible in describing such pathways can help biologists discover contradictory informations and guide them during experiments knowing how huge the cells metabolic networks have become. One of the extensions that can also be introduced is the notion of *Aboutness* [4] that can limit and focus search results to what seems relevant to a single or a group of entities (proteins).

**Acknowledgements:** This work is partially supported by the Région Midi-Pyrénées project called CLE, the Lebanese National Council for Scientific Research (LNCSR), and the French-Spanish lab LIRP. Comments from anonymous referees have largely contributed to the improvements of this paper.

## References

1. Hajnal Andréka, István Németi, and Johan van Benthem. Modal languages and bounded fragments of predicate logic. *Journal of Philosophical Logic*, 27(3):217–274, 1998.
2. Robert Demolombe. Syntactical characterization of a subset of domain-independent formulas. *J. ACM*, 39(1):71–94, 1992.
3. Robert Demolombe and Luis Fariñas del Cerro. An Inference Rule for Hypothesis Generation. In *Proc. of International Joint Conference on Artificial Intelligence*, Sydney, 1991.
4. Robert Demolombe and Luis Fariñas del Cerro. Information about a given entity: From semantics towards automated deduction. *J. Log. Comput.*, 20(6):1231–1250, 2010.
5. Robert Demolombe, Luis Fariñas del Cerro, and Naji Obeid. A logical model for metabolic networks with inhibition. *BIOCOMP'13*, 2013.
6. Martin Erwig and Eric Walkingshaw. Causal reasoning with neuron diagrams. In *Proceedings of the 2010 IEEE Symposium on Visual Languages and Human-Centric Computing*, VLHCC '10, pages 101–108, Washington, DC, USA, 2010. IEEE Computer Society.
7. V Glorian, G Maillot, S Poles, J S Iacovoni, G Favre, and S Vagner. Hur-dependent loading of mirna risc to the mrna encoding the ras-related small gtpase rhob controls its translation during uv-induced apoptosis. *Cell Death Differ*, 18(11):1692–70, 2011.
8. Katsumi Inoue. Linear resolution for consequence finding. *Artificial Intelligence*, 56(23):301 – 353, 1992.

9. Katsumi Inoue, Andrei Doncescu, and Hidetomo Nabeshima. Hypothesizing about causal networks with positive and negative effects by meta-level abduction. In *Proceedings of the 20th international conference on Inductive logic programming*, ILP'10, pages 114–129, Berlin, Heidelberg, 2011. Springer-Verlag.
10. Katsumi Inoue, Andrei Doncescu, and Hidetomo Nabeshima. Completing causal networks by meta-level abduction. *Machine Learning*, 91(2):239–277, 2013.
11. Kurt W Kohn and Yves Pommier. Molecular interaction map of the p53 and mdm2 logic elements, which control the off-on swith of p53 response to dna damage. *Biochem Biophys Res Commun*, 331(3):816–27, 2005.
12. Stephen Muggleton and Christopher H. Bryant. Theory completion using inverse entailment. In *Proceedings of the 10th International Conference on Inductive Logic Programming*, ILP '00, pages 130–146, London, UK, UK, 2000. Springer-Verlag.
13. Hidetomo Nabeshima, Koji Iwanuma, Katsumi Inoue, and Oliver Ray. Solar: An automated deduction system for consequence finding. *AI Commun.*, 23(2-3):183–203, April 2010.
14. Y. Pommier, O. Sordet, V.A. Rao, H. Zhang, and K.W. Kohn. Targeting chk2 kinase: molecular interaction maps and therapeutic rationale. *Curr Pharm Des*, 11(22):2855–72, 2005.
15. Oliver Ray. Automated abduction in scientific discovery. In *Model-Based Reasoning in Science and Medicine*, pages 103–116. Springer, June 2007.
16. Oliver Ray, Ken Whelan, and Ross King. Logic-based steady-state analysis and revision of metabolic networks with inhibition. In *Proceedings of the 2010 International Conference on Complex, Intelligent and Software Intensive Systems*, CISIS '10, pages 661–666, Washington, DC, USA, 2010. IEEE Computer Society.
17. Philip Reiser, Ross King, Douglas Kell, Stephen Muggleton, Christopher Bryant, and Stephen Oliver. Developing a logical model of yeast metabolism. *Electronic Transactions in Artificial Intelligence*, 5:233–244, 2001.
18. Raymond Reiter. Readings in nonmonotonic reasoning. chapter On closed world data bases, pages 300–310. Morgan Kaufmann Publishers Inc., San Francisco, CA, USA, 1987.
19. Joseph Shoenfield. *Mathematical logic*. Addison-Wesley series in logic. Addison-Wesley Pub. Co., 1967.
20. Pierre Siegel. *Representation et utilisation de la connaissance en calcul propositionnel*. Thèse d'État, Université d'Aix-Marseille II, Luminy, France, 1987.
21. Alireza Tamaddoni-Nezhad, Antonis C. Kakas, Stephen Muggleton, and Florencio Pazos. Modelling inhibition in metabolic pathways through abduction and induction. In Rui Camacho, Ross D. King, and Ashwin Srinivasan, editors, *Inductive Logic Programming, 14th International Conference, ILP 2004, Porto, Portugal, September 6-8, 2004, Proceedings*, volume 3194 of *Lecture Notes in Computer Science*, pages 305–322. Springer, 2004.
22. Jeffrey Ullman. *Principles of database systems*. Computer software engineering series. Computer Science Press, 1980.

# Translating the SBGN-AF language into logic to analyze signalling networks

Adrien Rougny[1,2,4] ⋆, Christine Froidevaux[3,4], Yoshitaka Yamamoto[5], and Katsumi Inoue[2]

[1] École Normale Supérieure de Lyon
15 parvis René Descartes, BP 7000 69342 Lyon, Cedex France
[2] National Institute of Informatics
2-1-2 Hitotsubashi, Chiyoda-ku, Tokyo, 101-8430, Japan
[3] Laboratoire de Recherche en Informatique (LRI), CNRS UMR 8623
Université Paris Sud, F-91405 Orsay, Cedex France
[4] AMIB INRIA
[5] Division of Medicine and Engineering Science
University of Yamanashi, 4-3-11 Takeda, Kofu, Yamanashi, 400-8511, Japan

**Abstract.** Systems Biology focuses on the understanding of complex biological systems as a whole. These systems are often represented in the literature as molecular networks, such as metabolic and signalling networks. With the rise of high-throughput data, automatic methods are needed in order to build and analyze networks that are bigger and bigger. Reasoning techniques seem to be suitable to perform these tasks for three reasons: (i) they do not need any quantitative biological parameters that are often hard to obtain, (ii) the processes that allowed to obtain the results are understandable by the biologist experts and can be explained and (iii) they allow to perform different tasks in the same formal framework. In this paper, we propose a translation into logics of the Systems Biology Graphical Notation Activity Flow language (SBGN-AF), which is a standard graphical notation used to represent molecular networks. We show how this translation can be used to analyze signalling networks with one analysis example.

## 1 Introduction

Systems Biology has emerged in early 2000's as a major field of biology and is now one of its most studied fields. Studying systems as a whole allows the understanding and the discovery of holistic biological properties and to gather and link pieces of knowledge that had until now remained independent. Systems Biology focuses on the study of different types of molecular networks, such as signalling or metabolic networks. A signalling network shows the transduction of a signal through a cell or a cell lineage, from the reception of the signal by the cell (such as the binding of an hormone to its receptor) to the cell's response (such as the expression of particular genes). For example, the *FSH-R signalling network* shows how the binding of FSH to its receptor triggers a signalling cascade that finally stimulates growth in female granulosa cells. As for a metabolic network, it shows the chain of reactions that occur in the cell that produce particular metabolites, and possibly how these reactions are regulated. For example, the *human glycolysis metabolic network* shows the chain of reactions that transforms glucose to pyruvate, producing energy in the form of ATP.

⋆ Electronic address: `rougny@lri.fr`; Corresponding author

Building and studying molecular networks is crucial to understand and to be able to change the cell's mechanisms and behavior. It is particularly important for healthcare: the precise understanding of the dynamics of the molecular mechanisms would allow us to design more efficient and side-effect-free drugs to cure particular diseases. With the appearance of high-throughput experiments, there has been an explosion of the quantity of experimental data available. Data are also more complex, as they come from different sources such as various experiment types. Therefore its integration in order to build molecular networks has become a challenge. Also, as the size of the networks increase more and more (the RECON 2 metabolic network has more than 2600 molecules and 7400 reactions [20]), new scalable methods to study the dynamics of the network are needed. Molecular networks have historically been (and still are) built and analyzed by hand by biologists. Since the 90's, automatic methods have arisen, but they have two drawbacks: (i) methods may focus on integrating only one type of experimental data [3][18][2], which runs counter to the holistic point of view and (ii) methods that analyze the dynamics of the network such as linear equations need quantitative parameters that are difficult to evaluate. Discrete reasoning techniques have been applied to build [1][6], refine [16] and analyze [21][4][17] molecular networks. They are suitable to perform these tasks for three reasons: (i) they do not use any quantitative parameters that are difficult to obtain, (ii) the processes that allowed to obtain the results are understandable by the biologist experts and can be explained and (iii) they allow to perform different tasks in the same formal framework.

The analyses range from the study of the topology of networks (reachability problems [10][19], steady-state analysis) to the study of the dynamics of networks (boolean network analysis, model checking of temporal properties [4]), as well as explaining observations or refining a network using experiments. All these analyses take as input and might give as ouput a molecular network formalized into logic-based formalisms (whether propositional logic, classical FOL, temporal logic or answer-set programming languages, etc). Molecular networks being usually found in the literature in a graphical form, a matching between graphical and logical representations of a network is needed.

In this paper, we propose a logical formalization of the Systems Biology Graphical Notation (SBGN) [12], which is a standard used to represent molecular networks, and especially metabolic and signalling networks. More precisely, we propose a translation of the different glyphs of SBGN-AF into first-order logic (FOL) so that any SBGN-AF network can be translated into FOL and logically deduced facts can be interpreted as elements of SBGN-AF networks. We then show that our translation can be used to analyze signalling networks. Section 2 introduces SBGN-AF and our translation, whereas Section 3 gives one example of signalling network analysis.

## 2 Translation of the SBGN-AF language to predicates and axioms

SBGN is a standard used to represent molecular networks. It is divided into three languages: the Process Diagram (SBGN-PD), the Entity Relationship diagram (SBGN-ER) and the Activity Flow diagram (SBGN-AF). Each language of SBGN has a different purpose and different features. Shortly, SBGN-PD is used to represent processes that make the state of biological objects (mainly of molecules) change, SBGN-ER to represent interaction rules between biological entities and SBGN-AF to represent influences of biological activities on each other. While SBGN-PD is usually used to represent metabolic networks, SBGN-AF is mainly used to represent signalling networks. SBGN has the following advantages: (i) its development is community-based, making it trustworthy and accessible to everyone, (ii) it is largely supported by Systems Biology softwares used to visualize, edit and analyze molecular networks, (iii) it is based on other standards such as the Systems Biology Ontology (SBO) [11].

A few translations of the SBGN languages have been proposed in various formalisms in order to store the networks (e.g. [22]) or to analyze them (e.g. [13]) but no translation from SBGN to logical formalisms have been yet proposed, although [5] proposes a translation of the MIM graphical notation, that can be considered as the "ancestor" of SBGN-ER to FOL. SBGN-AF is the most abstracted and simple language of SBGN. It is used to represent influences of biological activities on each other, and is therefore suitable to represent signalling network, where the understanding of the transduction of a signal is important. Translating SBGN-AF into logic allows to have a standard and unambiguous formalization of signalling networks into logic that is close to the biological semantics of the networks. It might lead to building more realistic, expressive and accurate models, which take into account the specificities of biological interactions.

### 2.1 Translation of the different glyphs of SBGN-AF

Each glyph of SBGN-AF (*i.e.* each drawn symbol that has a biological meaning, here nodes and arcs) is translated into one predicate. Unary predicates are used to translate the biological objects represented by nodes, while binary predicates are used to translate the biological relations represented by arcs. Each node will be associated to a constant used to name it. This constant might be the label associated to the node if it represents a biological object, or an arbitrary string if it is a logical operator. In the following translation, we use the variables $X$ and $Y$ for the sake of genericity. The translation to FOL of the different glyphs of SBGN-AF is given below. Each glyph is associated in [15] to an SBO term and a detailed explanation of its meaning.

## Activity

*Biological Activity*

Glyph:

Translation: $biologicalActivity(X) \equiv ba(X)$
Meaning: $X$ is a biological activity

*Phenotype*

Glyph:

Translation: $phenotype(X)$
Meaning: $X$ is a phenotype

*Perturbation*

Glyph:

Translation: $perturbation(X)$
Meaning: $X$ is a perturbation

## Modulation

*Positive Influence*

Glyph: 

Translation: $stimulates(X, Y)$
Meaning: $X$ has a positive influence on $Y$

*Negative Influence*

Glyph: 

Translation: $inhibits(X, Y)$
Meaning: $X$ has a negative influence on $Y$

*Unknown Influence*

Glyph: 

Translation: $unknownModulates(X, Y)$
Meaning: $X$ has an unknown influence on $Y$

*Necessary Stimulation*

Glyph: 

Translation: $necessaryStimulates(X, Y)$
Meaning: $X$ has to be performed for $Y$ to be performed

### Logical Operator

We introduce a new constant for each occurrence of a Logical Operator in the SBGN-AF network we want to translate. These constants name the Logical Operators and are needed to formalize the Logical Arcs.

*AND operator*

Glyph:

Translation: $andNode(X)$
Meaning: $X$ is an AND operator

*OR operator*

Glyph:

Translation: $orNode(X)$
Meaning: $X$ is an OR operator

*NOT operator*

Glyph:

Translation: $notNode(X)$
Meaning: $X$ is a NOT operator

*Delay operator*

Glyph:

Translation: $delayNode(X)$
Meaning: $X$ is a delay operator

*Logical Arc*

Glyph: 

Translation: $input(X, Y)$
Meaning: $X$ is an input of the Logical Operator Y

We may translate the AND and OR operators (together with their input and ouput arcs) in another way, as in [10]. Let $X$ be either an AND Operator, an OR Operator or an Activity, and $Y$ be the ouput of $X$ where $Y$ can be a Biological Activity or a Phenotype. We note $I(X)$ the set of inputs of $X$ if $X$ is a logical operator. Then the translation of the fact that "$X$ stimulates $Y$" is recursively obtained as follows:

- if $X$ is an Activity it is $stimulates(X, Y)$
- if $X$ is an OR Operator, it is the conjunction on each $S_i \in I(X)$ of the translation of "$S_i$ stimulates $Y$"

– if $X$ is an AND Operator, it is the disjunction on each $S_i \in I(X)$ of the translation of "$S_i$ stimulates $Y$"

Together with this translation we must introduce the unary predicate *present* and the axiom (1) that expresses the fact that, if "$X$ stimulates $Y$", $Y$ will be present if all the inputs of $X$ are present in the case where $X$ is an AND operator or at least one input of $X$ is present in the case where $X$ is an OR operator.

$$present(X) \wedge stimulates(X, Y) \Rightarrow present(Y) \quad (1)$$

This alternative translation is useful to avoid mapping the AND and OR operators to constants and to avoid using universal quantifiers in some axioms describing their behaviors.

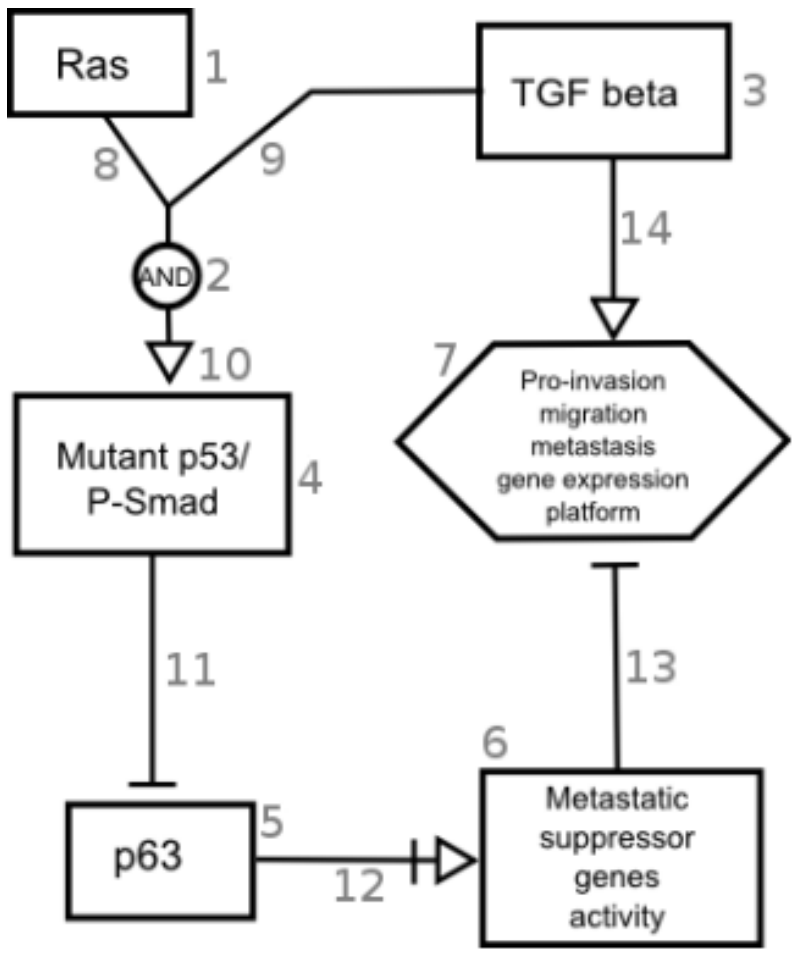

**Fig. 1.** Regulation of TGF-$\beta$-induced metastasis network

**An example of SBGN-AF network and its translation.** Figure 1, taken from [15], shows the simple *Regulation of TGF-$\beta$-induced metastasis* network represented in SBGN-AF. This network shows how the TGF-$\beta$ protein can induce cancerous cell proliferation. Following is the translation of the network.
1. $ba(ras)$ 2. $andNode(a1)$ 3. $ba(tgf_beta)$ 4. $ba(mut_p53_psmad)$ 5. $ba(p63)$
6. $ba(metastasis_suppressor)$ 7. $phenotype(metastasis)$ 8. $input(ras, a1)$
9. $input(tgf_beta, a1)$ 10. $stimulates(a1, mut_p53_psmad)$
11. $inhibits(mut_p53_psmad, p63)$
12. $necessaryStimulates(p63, metastasis_suppressor)$
13. $inhibits(metastasis_suppressor, metastasis)$
14. $stimulates(tgf_beta, metastasis)$

With the alternative translation of AND operators, the facts 2. 8. 9. 10. should be replaced by the disjunction

$$stimulates(ras, mut_p53_psmad) \vee stimulates(tgf_beta, mut_p53_psmad).$$

## 2.2 Ontological and Typing axioms

**Ontological axioms.** SBGN-AF contains three simple ontologies that can be extracted from SBO, the first one for the Activity Nodes, the second one for the Modulation Arcs and the last one for the Logical Operators. These ontologies are given in Fig. 2.

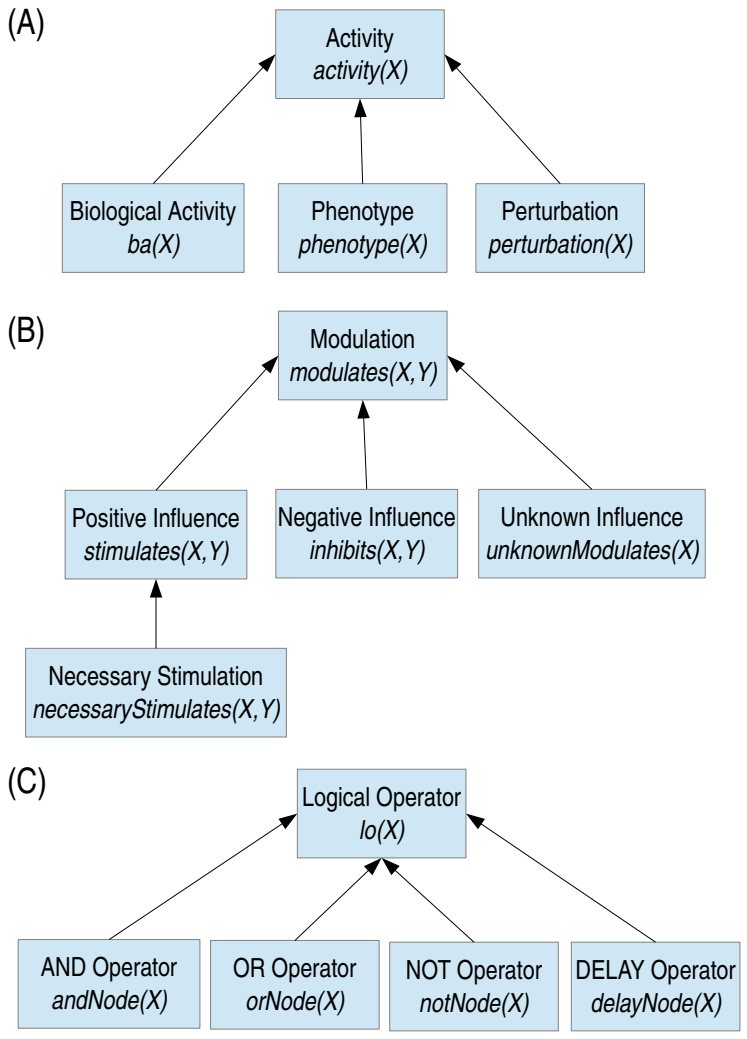

**Fig. 2.** The three ontologies of SBGN-AF. (A) Activity ontology. (B) Modulation ontology. (C) Logical Operator ontology. Boxes represent classes and arrows represent $is_a$ relations.

We introduce three new ontological predicates, one for each top class of the ontologies:

- $activity(X)$ means that $X$ is an Activity
- $modulates(X, Y)$ means that $X$ modulates $Y$
- $logicalOperator(X)$ (or $lo(X)$ for short) means that $X$ is a Logical Operator

Each class of any of the three ontologies has already been translated in Sec. 2.1. The binary relation $is_a$ is translated by the implication operator $\Rightarrow$. Then, for each ontology, for each relation of the type $subclass_i$ $is_a$ $superclass_j$, we will add the following axiom [1] to the theory:

$$subclass_i(X) \Rightarrow superclass_j(X) \tag{2}$$

For example, the axiom $ba(X) \Rightarrow activity(X)$ formalizes the fact that all Biological Activities are Activities. As, for all three ontologies, subclasses at the same level are disjoint, we will add for every pair of distinct subclasses $(subclass_i, subclass_j)$ of the same class the following constraint:

$$subclass_i(X) \wedge subclass_j(X) \Rightarrow \bot \tag{3}$$

For example, the axiom $phenotype(X) \wedge ba(X) \Rightarrow \bot$ formalizes the fact that a Biological Activity is not a Phenotype and conversely.

**Typing axioms.** Variables of binary predicates must be typed, as not all instantiations are allowed. The typing rules are obtained from the constraints of the language. For example, since a Negative Influence arc can have as input any Activity Node or Logical Operator and as ouput a Biological Activity or a Phenotype, the first variable of the predicate $inhibited/2$ can be instantiated by any constant that represents an Activity Node or a Logical Operator, whereas its second variable can be instantiated by any constant that represents either a Biological Activity or a Phenotype. Consequently, we add to our theory the following axiom:

$$inhibited(X_1, X_2) \Rightarrow (activity(X_1) \vee lo(X_1)) \wedge (ba(X_2) \vee phenotype(X_2)) \tag{4}$$

Together with the axioms describing the ontology and ontological facts, axiom (4) constrains the possible instantiations of the variables of the predicate $inhibited/2$.

## 3 Analyzing signalling networks using the translation

A standardized formalization of signalling networks using logic has to be accurate and useful for any kind of analysis and any type of reasoning technique, i.e. deduction, abduction, induction. We aim at providing a logic-based translation method that would be relevant for any signalling network and that would give a formal basis for any further formal analyis. As an example of such an analysis, we take the boolean analysis.

**Boolean network analysis.** Boolean analysis is widely used to study the dynamics [7][4] of signalling network or the transduction of the signal [5] in such networks. We propose here a set of general axioms based on our translation that allow the study of the dynamics of any SBGN-AF signalling network. This kind of axioms can be found in [7] and in [5] in first-order. In [7], the authors propose a general translation of Boolean networks into ASP and study their dynamics when updated synchronously, whereas in [5], the authors propose a translation of the MIM gaphical notation [14] into FOL as well as a set of axioms to study the transduction of the signal in signalling networks. Moreover, they propose a method to abduce explanations about the activation/inhibition of a given molecule. Whereas our language is richer than the one used in [7] who has only two types of arcs, it is complementary to the one used in [5] while overlapping. Our set of axioms is based on the following choices of modeling, that are not found in [5] or [7]:

- inhibition is preferred to stimulation, meaning that if an Activity is at the same time stimulated and inhibited, it will be not be performed [10][8]
- an Activity that has no stimulator and at least one inhibitor will be performed if it is not inhibited.

[1] In this axiom and for the rest of the article, universal quantifiers are implicit. They are omitted for the sake of readability.

Also, unlike the languages in [5] and [7] our language includes Necessary Stimulations and AND operators.
Now we give the axioms expressing the dynamics of the network. We introduce the predicate $present/2$: if $X$ is a Biological Activity, $present(X,T)$ means that $X$ is performed at time $T$. For the sake of concision, we use the predicate $present/2$ with Logical Operators as arguments even if this has no biological meaning. Following are the axioms needed to study the dynamics of any SBGN-AF signalling network. We formulate them in natural language and give their translation into FOL:

- An Activity $X$ that has at least one modulator is present at time $T+1$ if the following conditions are satisfied:
  - none of its inhibitors is present at time $T$
  - all its necessary stimulators are present at time $T$
  - in the case it has at least one stimulator, one of them is present at time $T$

$$\begin{aligned} activity(X) \wedge \{&\exists M[modulates(M,X)]\} \\ &\wedge \{\forall I[inhibits(I,X) \Rightarrow \neg present(I,T)]\} \\ &\wedge \{\forall N[necessarlyStimulates(N,X) \Rightarrow present(N,T)]\} \\ &\wedge \{(\exists S[stimulates(S,X)]) \Rightarrow \exists S'[stimulates(S',X) \\ &\qquad\qquad\qquad\qquad \wedge present(S',T)]\} \\ &\Rightarrow present(X,T+1) \end{aligned} \tag{5}$$

- If an Activity $X$ has no modulator and is present at time $T$ then it is present at time $T+1$:

$$\begin{aligned} activity(X) \wedge &\neg\exists M[modulates(M,X)] \\ &\wedge present(X,T) \\ &\Rightarrow present(X,T+1) \end{aligned} \tag{6}$$

– An AND Operator $A$ is present at time $T$ if all its inputs are present at time $T$:

$$andNode(A) \wedge \forall J[input(J, A) \Rightarrow present(J, T)] \Rightarrow present(A, T) \quad (7)$$

– An OR Operator $O$ is present at time $T$ if at least one of its inputs is present at time $T$:

$$orNode(O) \wedge \exists J[input(J, O) \wedge present(J, T)] \Rightarrow present(O, T) \quad (8)$$

The dynamics, and particularly the steady states (or attractor points) can be deduced from the above axioms in different ways, that all use non-mononotonic reasoning. In [7], the authors compute all the state transitions using Answer Set Programming (ASP) by incrementing the temporal parameter $T$ and detect afterwards the steady-states. This method is costly as it is necessary to compute all state transitions but it allows to get the full dynamics of the network. Another method consists in eliminating the quantifiers of the theory by completing the formulas under the Closed World Assumption (CWA), as in [5]. Then the steady states are the models of the resulting theory. In [9], the authors show that Boolean networks are equivalent to Normal Logic Programs (NLP) and that the steady states of such a network are the supported models of its corresponding NLP.

**Steady states of our example.** We instantiate our axioms and simplify them with the facts obtained from the translation of our SBGN-AF network example in Sec. 2.1. We apply the CWA: we consider that the only modulations are those given by the network translation. Since we want to deduce the steady states, the time parameter $T$ is no longer needed. We obtain the following rules:

$$present(ras) \Rightarrow present(ras) \quad (9)$$

$$present(tgf_beta) \Rightarrow present(tgf_beta) \quad (10)$$

$$present(ras) \wedge present(tgf_beta) \Rightarrow present(a1) \quad (11)$$

$$present(a1) \Rightarrow present(mut_p53_psmad) \quad (12)$$

$$\neg present(mut_p53_psmad) \Rightarrow present(p63) \quad (13)$$

$$present(p63) \Rightarrow present(suppressor) \quad (14)$$

$$present(tgf_beta) \wedge \neg present(suppressor) \Rightarrow present(metastasis) \quad (15)$$

Rules (9) and (10) are obtained from axiom (6), rule (11) from axiom (7) whereas the other ones are obtained from axiom (5). Computing the supported models of rules (9-15) gives four steady states depending on the initial presence of *RAS* and *TGF-β*:

– if either *RAS* or *TGF-β* is absent at the initial state, then *p63* and the *metastatic suppressor* will be present at steady-state, making the *metastasis* absent.
– if *RAS* and *TGF-β* are present at the initial state, then *p63* and the *metastatic suppressor* will be absent at steady-state, making the *metastasis* present.

## 4 Conclusion

We have proposed a translation of the SBGN-AF into FOL. This translation allows to formalize signalling networks into logic in a standardized and meaningful way regarding Systems Biology. We showed with one short example that axioms used to analyze signalling networks could be expressed in this formalism. We plan to investigate whether we can perform any type of analysis within this formalism. Another research direction will be the search for other relevant axioms than the typing and ontological ones that are independent of the modeling choices and intrisic to SBGN-AF.

## References

1. Z. Aslaoui-Errafi, S. Cohen-Boulakia, C. Froidevaux, P. Gloaguen, A. Poupon, A. Rougny, and M. Yahiaoui. Towards a logic-based method to infer provenance-aware molecular networks. In *Proc. of the 1st ECML/PKDD International workshop on Learning and Discovery in Symbolic Systems Biology (LDSSB)*, pages 103–110, Bristol, Royaume-Uni, Sept. 2012.
2. M. Bansal, V. Belcastro, A. Ambesi-Impiombato, and D. Di Bernardo. How to infer gene networks from expression profiles. *Molecular systems biology*, 3(78), 2007.
3. T. Çakır, M. M. Hendriks, J. A. Westerhuis, and A. K. Smilde. Metabolic network discovery through reverse engineering of metabolome data. *Metabolomics*, 5(3):318–329, 2009.
4. L. Calzone, F. Fages, and S. Soliman. Biocham: an environment for modeling biological systems and formalizing experimental knowledge. *Bioinformatics*, 22(14):1805–1807, 2006.
5. R. Demolombe, L. F. del Cerro, and N. Obeid. A logical model for metabolic networks with inhibition. *14th International Conference on Bioinformatics and Computational Biology, In Print*, 2013.
6. F. Eduati, A. Corradin, B. Di Camillo, and G. Toffolo. A boolean approach to linear prediction for signaling network modeling. *PLoS One*, 5(9):e12789, 2010.
7. T. Fayruzov, J. Janssen, D. Vermeir, and C. Cornelis. Modelling gene and protein regulatory networks with answer set programming. *International journal of data mining and bioinformatics*, 5(2):209–229, 2011.
8. I. Gat-Viks and R. Shamir. Chain functions and scoring functions in genetic networks. *Bioinformatics*, 19(suppl 1):i108–i117, 2003.
9. K. Inoue. Logic programming for boolean networks. In T. Walsh, editor, *IJCAI*, pages 924–930. IJCAI/AAAI, 2011.
10. K. Inoue, A. Doncescu, and H. Nabeshima. Completing causal networks by meta-level abduction. *Machine Learning*, 91(2):239–277, 2013.
11. N. Le Novère. Model storage, exchange and integration. *BMC neuroscience*, 7(Suppl 1):S11, 2006.
12. N. Le Novere et al. The systems biology graphical notation. *Nature biotechnology*, 27(8):735–741, 2009.
13. L. Loewe, M. Guerriero, S. Watterson, S. Moodie, P. Ghazal, and J. Hillston. Translation from the quantified implicit process flow abstraction in sbgn-pd diagrams to bio-pepa illustrated on the cholesterol pathway. 6575:13–38, 2011.
14. A. Luna, E. Karac, M. Sunshine, L. Chang, R. Nussinov, M. Aladjem, and K. Kohn. A formal mim specification and tools for the common exchange of mim diagrams: an xml-based format, an api, and a validation method. *BMC bioinformatics*, 12(1):167, 2011.

15. H. Mi, F. Schreiber, N. Le Novére, S. Moodie, and A. Sorokin. Systems biology graphical notation: activity flow language level 1. *Nature Precedings*, 713, 2009.
16. M. K. Morris, J. Saez-Rodriguez, D. C. Clarke, P. K. Sorger, and D. A. Lauffenburger. Training signaling pathway maps to biochemical data with constrained fuzzy logic: quantitative analysis of liver cell responses to inflammatory stimuli. *PLoS computational biology*, 7(3):e1001099, 2011.
17. M. K. Morris, J. Saez-Rodriguez, P. K. Sorger, and D. A. Lauffenburger. Logic-based models for the analysis of cell signaling networks. *Biochemistry*, 49(15):3216–3224, 2010.
18. K. Sachs, O. Perez, D. Pe'er, D. A. Lauffenburger, and G. P. Nolan. Causal protein-signaling networks derived from multiparameter single-cell data. *Science*, 308(5721):523–529, 2005.
19. T. Soh, K. Inoue, T. Baba, T. Takada, and T. Shiroishi. Evaluation of the prediction of gene knockout effects by minimal pathway enumeration. *International Journal On Advances in Life Sciences*, 4(3 and 4):154–165, 2012.
20. I. Thiele et al. A community-driven global reconstruction of human metabolism. *Nature biotechnology*, 31(5):419–425, 2013.
21. A. Tiwari, C. Talcott, M. Knapp, P. Lincoln, and K. Laderoute. Analyzing pathways using sat-based approaches. In *Algebraic biology*, pages 155–169. Springer, 2007.
22. M. P. van Iersel et al. Software support for sbgn maps: Sbgn-ml and libsbgn. *Bioinformatics*, 28(15):2016–2021, 2012.

# Practically Fast Non-monotone Dualization based on Monotone Dualization

Yoshitaka Yamamoto, Koji Iwanuma, Nabeshima Hidetomo

University of Yamanashi
4-3-11 Takeda, Kofu-shi, Yamanashi 400-8511, Japan.

**Abstract.** This paper investigates practically fast methods for dualizing non-monotone Boolean functions. It is known that non-monotone dualization (NMD) can be logically reduced into two equivalent monotone dualization (MD) tasks. This reduction enables us to solve the original non-monotone dualization problem by the state-of-the art MD methods that have been extensively studied so far. In this paper, we focus on two MD methods: the one is based on so-called *enumeration tree*, and the other is based on BDD and Zero-suppressed BDD (ZDD). We first propose an approximation algorithm using the MD method based on enumeration trees. This algorithm removes redundant clauses though the MD computation, and thus is used as a pre-filter in NMD. We secondly propose a new algorithm using MD method based on BDD/ZDDs.

## 1 Introduction

The problem of *non-monotone dualization* (NMD) is to generate an irredundant prime CNF formula $\psi$ of the dual $f^d$ where $f$ is a *general* Boolean function represented by CNF [1]. The DNF formula $\phi$ of $f^d$ is easily obtained by De Morgan's laws interchanging the connectives of the CNF formula. Hence, the main task of NMD is to convert the DNF $\phi$ to an equivalent CNF $\psi$.

This translation is often seen in finding an alternative representation of the input form. For instance, given a set of models, it might be desirable to seek underlying structure behind the models. In contrast, by converting a CNF formula into DNF, we obtain the models satisfying the CNF formula, which is useful for model checking in discrete systems. This fact shows an easy reduction from SAT problems to NMD, and conjectures its high complexity. The research thus has been much focused on some restricted classes of Boolean functions.

*Monotone dualization* (MD) is one such class that deals with *monotone* Boolean functions for which CNF formulas are negation-free [2]. MD is one of the few problems whose tractability status with respect to polynomial-total time is still unknown. Besides, it is known that MD has many equivalent problems in discrete mathematics, such as the minimal hitting set enumeration and the hypergraph transversal computation [2].

MD class has received much attention that yields remarkable algorithms. In terms of the complexity, MD is solvable in a quasi-polynomial-total time [3]. In the practical point of view, there are two fast MD algorithms [4–6]. The first one [4, 5] is based on the *reverse search* [7]. This algorithm uses so-called *parent-child relationship* over the solutions to structure the search space as a rooted-tree. This is called an *enumeration tree*. Then, it searches the enumeration tree for the solutions in the depth-first search manner. The second one [6] uses the Binary Decision Diagram (BDD) and the Zero-suppressed BDD (ZDD) that are the compressed data structures for set families. This MD computation is achieved by recursively applying family algebra operations in BDD/ZDDs.

In this paper, we investigate practically fast methods for NMD based on two state-of-the-art MD algorithms. We can apply MD to the original problem by handling negated variables as regular variables. However, the output of MD can contain *redundant clauses*, like resolvents and tautologies, to be removed. In terms of this drawback of NMD, the literature [8] shows that adding some *tautologies* can prohibit from generating resolvents through MD computation.

Using the deterrent effect by adding tautologies, we first propose an approximation algorithm. It can output a CNF formula consisting of prime and *mostly* irredundant implicates by one time MD computation. In other words, this is regarded as a “pre-filtering” tool before obtaining the NMD output. The performance is reported using randomly generated instances in the paper. We next propose a new NMD algorithm based on the MD method with BDD/ZDD. This algorithm consists of two steps: it first computes the MD output and next minimizes the NMD output in BDD and ZDD. This paper shows a preliminary result on the scalability obtained by several inductive learning problems.

It should be emphasized that developing practically fact NMD is important for learning and non-monotonic reasoning. Especially, the task of NMD is often seen in inductive reasoning. In *learning from entailment*, there are several ILP methods [9, 10] that require NMD computation. In turn, *learning from interpretation* [11] uses NMD to derive formulas from the given models. In *brave induction* [12], which is an extended formalization of *learning from entailment*, the original procedure in the literature also requires NMD computation (i.e., CNF-DNF translation). Even if we only treat Horn programs, we require NMD, instead of MD, in those setting of inductive learning. Accordingly, if the learned programs are represented in non-Horn or normal formalization, NMD is also necessary.

The rest of this paper is organized as follows. Section 2 describes the notion and terminologies in the paper, and introduces the reduction technique from NMD to MD. Section 3 and 4 propose two approximation algorithms with enumeration trees and the BDD/ZDDs, respectively. Section 5 then concludes.

# 2 Background

## 2.1 Preliminaries

A *Boolean function* is a mapping $f : \{0,1\}^n \to \{0,1\}$. We write $g \models f$ if $f$ and $g$ satisfy $g(v) \leq f(v)$ for all $v \in \{0,1\}^n$. $g$ is (*logically*) *equivalent* to $f$, denoted by $g \equiv f$, if $g \models f$ and $f \models g$. A function $f$ is *monotone* if $v \leq w$ implies $f(v) \leq f(w)$ for all $v, w \in \{0,1\}^n$; otherwise it is *non-monotone*. Boolean variables $x_1, x_2, \ldots, x_n$ and their negations $\overline{x_1}, \overline{x_2}, \ldots, \overline{x_n}$ are called *literals*. The *dual* of a function $f$, denoted by $f^d$, is defined as $\overline{f}(\overline{x})$ where $\overline{f}$ and $\overline{x}$ is the negation of $f$ and $x$, respectively.

A *clause* (*resp. term*) is a disjunction (*resp.* conjunction) of literals which is often identified with the set of its literals. It is known that a clause is *tautology* if it contains complementary literals. A clause $C$ is an *implicate* of a function $f$ if $f \models C$. An implicate $C$ is *prime* if there is no implicate $C'$ such that $C' \subset C$.

A *conjunctive normal form* (CNF) (*resp. disjunctive normal form* (DNF)) formula is a conjunction of clauses (*resp.* disjunction of terms) which is often identified with the set of clauses in it. A CNF formula $\phi$ is *irredundant* if $\phi \not\equiv \phi - \{C\}$ for every clause $C$ in $\phi$; otherwise it is *redundant*. $\phi$ is *prime* if every clause in $\phi$ is a prime implicate of $\phi$; otherwise it is *non-prime*. Let $\phi_1$ and $\phi_2$ be two CNF formulas. $\phi_1$ *subsumes* $\phi_2$, denoted by $\phi_1 \succeq \phi_2$, if there is a clause $C \in \phi_1$ such that $C \subseteq D$ for each clause $D \in \phi_2$. In turn, $\phi_1$ *minimally subsumes* $\phi_2$, denoted by $\phi_1 \succeq^\natural \phi_2$, if $\phi_1$ subsumes $\phi_2$ but $\phi_1 - \{C\}$ does not subsume $\phi_2$ for every clause $C \in \phi_1$.

Let $\phi$ be a CNF formula. $\tau(\phi)$ denotes the CNF formula obtained by removing all tautologies from $\phi$. We say $\phi$ is *tautology-free* if $\phi = \tau(\phi)$. Now, we formally define the dualization problem as follows.

**Definition 1 (Dualization problem).**

| | |
|---|---|
| **Input**: | A tautology-free CNF formula $\phi$ |
| **Output**: | An irredundant prime CNF formula $\psi$ such that $\psi$ is logically equivalent to $\phi^d$ |

We call it *monotone dualization* (MD) if $\phi$ is negation-free; otherwise it is called *non-monotone dualization* (NMD). As well known [2], the task of MD is equivalent to enumerating the *minimal hitting sets* (MHSs) of a family of sets.

## 2.2 MD as MHS enumeration

**Definition 2 ((Minimal) Hitting set).** Let $\Pi$ be a finite set and $\mathcal{F}$ be a subset family of $\Pi$. A finite set $E$ is a *hitting set* of $\mathcal{F}$ if for every $F \in \mathcal{F}$, $E \cap F \neq \emptyset$. A finite set $E$ is a *minimal hitting set* (MHS) of $\mathcal{F}$ if $E$ satisfies that

1. $E$ is a hitting set of $\mathcal{F}$;
2. For every subset $E' \subseteq E$, if $E'$ is a hitting set of $\mathcal{F}$, then $E' = E$.

Note that any CNF formula $\phi$ can be identified with the family of clauses in $\phi$. Now, we consider the CNF formula, denoted by $M(\phi)$, which is the conjunction of all the MHSs of the family $\phi$. Then, the following holds.

**Theorem 1.** [13] *Let $\phi$ be a tautology-free CNF formula. A clause $C$ is in $\tau(M(\phi))$ if and only if $C$ is a non-tautological prime implicate of $\phi^d$.*

Hence, the output of MD for $\phi$ uniquely corresponds to $\tau(M(\phi))$: the set of all MHSs of the family $\phi$ by Theorem 1.

### 2.3 NMD as MHS enumeration

While MD is done by the state-of-the-art algorithms to compute MHSs [4–6], it is not straightforward to directly use them for NMD. The drawback lies in the appearance of redundant clauses. Hence, $\tau(M(\phi))$ is prime but not irredundant.

*Example 1.* Let the input CNF formula $\phi$ be $\{\{x_1,\ \overline{x_2},\ \overline{x_3}\},\ \{\overline{x_1},\ x_2,\ x_3\}\}$. $\tau(M(\phi))$ consists of the non-tautological prime implicates as follows:

$$\tau(M(\phi)) = \{\{x_1,\ x_2\},\ \{\overline{x_1},\ \overline{x_3}\},\ \{\overline{x_2},\ x_3\},\ \{x_1,\ x_3\},\ \{\overline{x_1},\ \overline{x_2}\},\ \{\overline{x_3},\ x_2\}\}.$$

We may notice that there are at least two irredundant subsets of $\tau(M(\phi))$:

$$\psi_1 = \{\{x_1,\ x_2\},\ \{\overline{x_1},\ \overline{x_3}\},\ \{\overline{x_2},\ x_3\}\}.\ \psi_2 = \{\{x_1,\ x_3\},\ \{\overline{x_1},\ \overline{x_2}\},\ \{\overline{x_3},\ x_2\}\}.$$

Indeed, $\psi_1$ is equivalent to $\psi_2$, and thus both are equivalent to $\tau(M(\phi))$.

Once $\tau(M(\phi))$ is obtained by MD computation, we need to *minimize* $\tau(M(\phi))$ so that it does not contain redundant clauses. In terms of the minimization problem, there are well-established tools such as *Espresso* [14]. There is also a well-known algorithm called *Minato-Morreale* algorithm [15] in the context of BDD/ZDD. In this paper, we focus on an alternative approach [8] that uses the following formula obtained by adding *tautologies* to the original formula.

**Definition 3 (Bottom formula).** Let $\phi$ be a tautology-free CNF formula and $Taut(\phi)$ be $\{\ x \vee \overline{x}\ \mid\ \phi \textit{ contains both } x \textit{ and } \overline{x}\ \}$. Then, the *bottom formula wrt* $\phi$ (in short, bottom formula) is defined as the CNF formula $\tau(M(\phi \cup Taut(\phi)))$.

### 2.4 Properties of bottom formulas

Bottom formulas have interesting properties [8].

**Theorem 2.** [8] *Let $\phi$ be a tautology-free CNF formula. Then, the bottom formula wrt $\phi$ is irredundant.*

*Example 2.* Recall the CNF formula $\phi$ in Example 1. Since $Taut(\phi)$ is the set $\{\{x_1,\ \overline{x_1}\},\ \{x_2,\ \overline{x_2}\},\ \{x_3,\ \overline{x_3}\}\}$, the bottom formula is as follows:

$$\{\{x_1, x_2, x_3\},\ \{\overline{x_3}, x_2, x_1\},\ \{\overline{x_3}, x_2, \overline{x_1}\},\ \{\overline{x_2}, x_3, x_1\},\ \{\overline{x_2}, x_3, \overline{x_1}\},\ \{\overline{x_2}, \overline{x_3}, \overline{x_1}\}\}.$$

We write $C_1, C_2, \ldots, C_6$ for the above clauses in turn (i.e., $C_4$ is $\{\overline{x_2},\ x_3,\ x_1\}$). We then notice that the bottom formula is irredundant, because every clause cannot be derived from the other ones. However, it is non-prime, since it contains a non-prime implicate $C_1$ whose subset $\{x_1,\ x_2\}$ is an implicate of $\phi^d$.

Every NMD output is logically described with the bottom formula as follows.

**Theorem 3.** [8] Let $\phi$ be a tautology-free CNF formula. $\psi$ is an NMD output of $\phi$ iff $\psi \subseteq \tau(M(\phi))$ and $\psi$ minimally subsumes the bottom formula wrt $\phi$.

*Example 3.* Recall Example 1 and Example 2. Fig. 1 describes the subsumption lattice bounded by two irredundant prime outputs $\psi_1$ and $\psi_2$ as well as the bottom formula $\{C_1, C_2, \dots, C_6\}$. The solid (*resp.* dotted) lines show the subsumption relation between $\psi_1$ (*resp.* $\psi_2$) and the bottom formula. We then notice that both outputs $\psi_1$ and $\psi_2$ minimally subsume the bottom formula.

By Theorem 3, any clause $C$ in $\tau(M(\phi))$ can be removed if $\tau(M(\phi)) - \{C\}$ subsumes the bottom theory. Hence, minimization of $\tau(M(\phi))$ is reduced to the set covering problem in $\tau(M(\phi))$ wrt the bottom theory. Based on this notion, we consider practically fast NMD using two state-of-the-art MD algorithms.

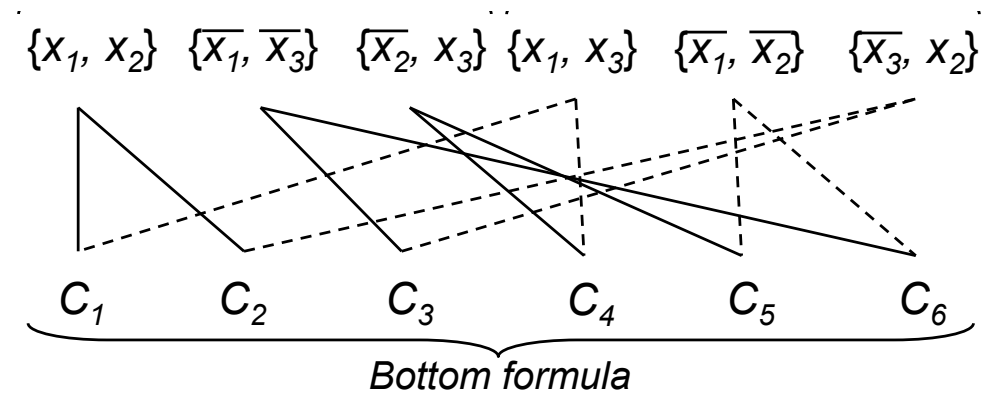

**Fig. 1.** Subsumption lattice bounded by NMD outputs and the bottom formula

# 3 NMD with enumeration trees

## 3.1 A practically fast MD algorithm with enumeration trees

Here, we focus on the fast algorithm for computing $\tau(M(\phi))$ by reverse search [4, 5]. This algorithm uses so-called *parent-child relationship* over the solutions:

**Definition 4 (Parent-child relationship [4]).** Let $\phi_i = \{C_1, \dots, C_i\}$ $(1 \le i \le n)$ be a subset family of $i$ clauses and $E_i$ be a minimal hitting set of $\phi_i$. Then a pair $(i+1, E_{i+1})$ is a *child* of a pair $(i, E_i)$ if $E_{i+1}$ satisfies the following:

- If $E_i$ is a minimal hitting set of $\phi_{i+1}$, then $E_{i+1} = E_i$.
- Else, $E_{i+1} = E_i \cup \{e\}$, where $e$ is an element in $\phi_{i+1}$ such that $E_i \cup \{e\}$ is a minimal hitting set of $\phi_{i+1}$.

This acyclic relationship structures the search space as a rooted-tree, called an *enumeration tree*. Using the enumeration tree, the algorithm searches for solutions (i.e., the non-tautological minimal hitting sets of $\phi$) in the depth-first search manner. Fig. 2[1] sketches it briefly [16].

In other words, this algorithm incrementally searches for an MHS of the next family $\phi_{i+1}$ from the MHS of the current family $\phi_i$. Uno shows [4] that its average computation for randomly generated instances is experimentally $O(n)$ per output, where $n$ is the input size.

[1] Since the original version is used for computing $M(\phi)$, we modify it so as to remove the tautologies of $M(\phi)$ by way of (1) in Fig. 2.

**Global** $\phi_n = \{C_1, \ldots, C_n\}$
*compute*($i$, $mhs$, $S$)
/*$mhs$ is an MHS of $\phi_i$ $(1 \leq i \leq n)$. $S$ is the family of MHSs of $\phi_n$.*/
**Begin**
if i = n then add $mhs$ to $S$ and return;
else if $mhs$ is an MHS of $\phi_{i+1}$ then do *compute*($i+1$, $mhs$, $S$);
else $\forall e \in C_{i+1}$ s.t. $mhs \cup \{e\}$ is a non tautological MHS of $\phi_{i+1}$ (1)
    do *compute*($i+1$, $mhs \cup \{e\}$, $S$);
output $S$ and return;
**End**

**Fig. 2.** A practically fast algorithm for computing $\tau(M(\phi_n))$

### 3.2 A practically fast pre-filtering algorithm: CoCo1

Based on the above MD algorithm, we propose a practically fast pre-filtering algorithm, named CoCo1, which removes redundant prime implicates in advance at the first step computing the bottom formula. CoCo1 works well in case that the size of bottom formula becomes exponentially large. The key idea lies in that we *approximate* the subsumption lattice upper bounded by the prime implicates and lower bounded by the bottom formula as the enumeration tree.

*Example 4.* Recall the CNF $\phi_5$ in Example 2 as follows:

$$\phi_5 = \{\{x_1, \overline{x_2}, \overline{x_3}\},\ \{\overline{x_1}, x_2, x_3\},\ \{x_1, \overline{x_1}\},\ \{x_2, \overline{x_2}\},\ \{x_3, \overline{x_3}\}\}.$$

Fig. 3 describes the enumeration tree for $\phi_5$ and captures the relation between $\tau(M(\phi))$ and the bottom theory wrt $\phi$.

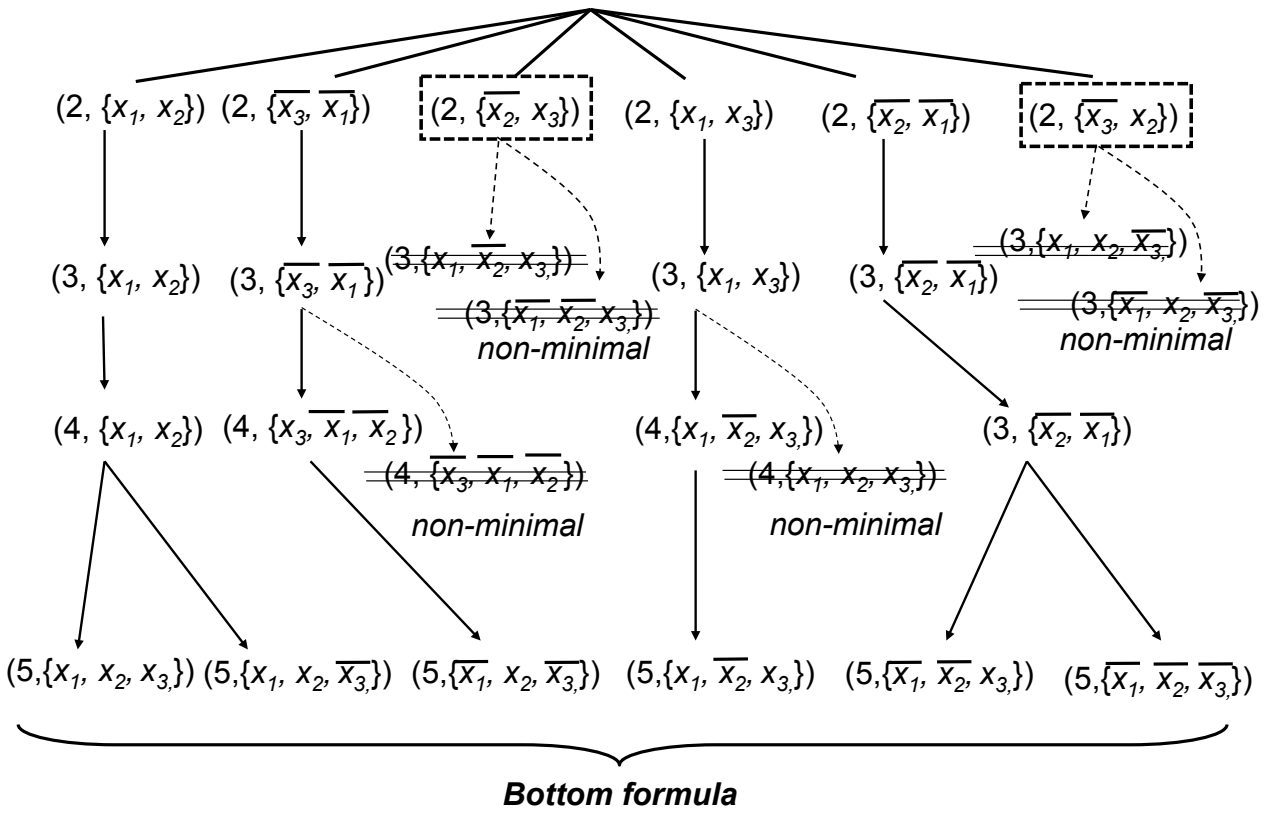

**Fig. 3.** Enumeration tree for $\phi_5$ from $\tau(M(\phi_2))$

We then notice that two prime implicates $\{\overline{x_2},\ x_3\}$ and $\{\overline{x_3},\ x_2\}$ do not involve generating the bottom formula. Indeed, every clause in the bottom formula is connected with another prime implicate. In other words, even if these two clauses are eliminated from the prime implicates, the remained ones subsume the bottom formula. Based on this observation, we establish a pre-filtering algorithm so that any prime implicate is removed unless there is a path from it to some clause of the bottom formula over the enumeration tree.

**Global** $\phi_n = \{C_1, \ldots, C_n\}$
*compute*($i$, $m$, $mhs$, $S$, $C$)
/* $m$ is the number of tautologies in $\phi_n$. $mhs$ is a MHS of $\phi_i$ ($1 \leq i \leq n$).
$S$ is a subset of the MHSs of $\phi_{n-m}$. $C$ is the copy of a MHS of $\phi_{n-m}$. */
**Begin**
if $i = n - m$, then copy $mhs$ to $C$;
if $i = n$, then add $C$ to $S$ and return false;
else if $mhs$ is a MHS of $\phi_{i+1}$
  return *compute*($i + 1$, $m$, $mhs$, $S$, $C$);
else $\forall e \in C_{i+1}$ s.t. $mhs \cup \{e\}$ is a non tautological MHS of $\phi_{i+1}$, do
  cont := *compute*($i + 1$, $m$, $mhs \cup \{e\}$, $S$, $C$);
  if cont = false and $i > n - m$ then return false;
return true;
**End**

**Fig. 4.** CoCo1 algorithm

Fig. 4 sketches the algorithm in brief. We name it CoCo1, which comes from the feature that it only **co**nfirms **one co**nnection between the prime implicates and the bottom formula over the enumeration tree. On the other hand, CoCo1 is not necessarily complete for removing every redundant prime implicate. Because the paths in the enumeration tree cannot correspond to the subsumption lattice. Indeed, there are several subsumption connections in Fig. 1 that disappear in Fig. 3. Despite that, CoCo1 can eliminate most of redundant clauses.

We have implemented CoCo1 and evaluated its performance by experiments. Each problem is randomly generated with the three parameters: The number of variables $E \in \{10,\ 20\}$, the size of clauses $N \in \{10,\ 30,\ 50\}$ and the average probability $P = 60$ that each variable appears in each clause. Table 1 describes the performance of MHS computation without CoCo1 filtering (i.e. baseline algorithm computing all the non-tautological prime implicates in Fig. 2) and with it, respectively, in terms of the size of the output, the CPU time $T$ (msec) and the number $R$ of redundant clauses contained in the output. For instance, in the problem of $P = 60$, $N = 30$ and $E = 20$, we can confirm that it computes totally 108094 prime implicates in 140 (msec) but 105586 of them are redundant.

Note that we use MiniSAT solver for checking whether or not each clause $C$ of the output $S$ is redundant in such a way that the solver checks the satisfiability of $(S - \{C\}) \wedge \overline{C}$ within (totally) 10 minutes. According to Table 1, we notice that CoCo1 can eliminate 90% of the redundant prime implicates on average, while the executing time is not so different from each other.

On the other hand, we notice that redundant clauses still remain in the output of CoCo1. It would be a considerable way to use some efficient minimization tools for eliminating those remained redundant clauses. We have used

*Espresso* [14] that is a well-known software for minimizing logical circuits. Thus, we now consider the two-phased NMD algorithm that first computes the MD output which can contain redundant clauses and then compute an NMD output by eliminating those redundant clauses using Espresso. Note that Espresso cannot directly derive the NMD output, since Espresso only minimizes the original formula, not translating it into another form. We have used the baseline MD algorithm and CoCo1 for the first step. Table 2 describes the result by applying Espresso to the outputs of the baseline algorithm and CoCo1, respectively. Note here that we have limited the executing time by 10 minutes. We remark that Espresso can work in the problems of $E = 20$, since most of redundant clauses are eliminated by CoCo1 in advance.

| Instances | | Baseline | | | CoCo1 | | |
|---|---|---|---|---|---|---|---|
| $P = 60$ | Perf. | $N = 10$ | $N = 30$ | $N = 50$ | $N = 10$ | $N = 30$ | $N = 50$ |
| | Size | 181 | 382 | 300 | 60 | 131 | 148 |
| $E = 10$ | R | 142 | 295 | 207 | 21 | 52 | 53 |
| | T (msec) | 14 | 21 | 26 | 19 | 24 | 29 |
| | Size | 11593 | 108094 | 322982 | 262 | 2200 | 3459 |
| $E = 20$ | R | 11495 | 105586 | T.O. | 162 | 1722 | 509 |
| | T (msec) | 6.1 | 140 | 640 | 12 | 220 | 870 |

**Table 1.** MHS computation with CoCo1

| Instances | | Baseline & Espresso | | | CoCo1 & Espresso | | |
|---|---|---|---|---|---|---|---|
| $P = 60$ | Perf. | $N = 10$ | $N = 30$ | $N = 50$ | $N = 10$ | $N = 30$ | $N = 50$ |
| $E = 10$ | Size | 31 | 45 | 69 | 33 | 47 | 71 |
| | T (msec) | 24 | 36 | 37 | 25 | 33 | 39 |
| $E = 20$ | Size | | | | 42 | 154 | 330 |
| | T (msec) | T.O. | T.O. | T.O. | 17 | 750 | 19260 |

**Table 2.** Espresso and MHS computation with CoCo1

## 4 NMD with BDD/ZDD

Here, we investigate NMD using the BDD/ZDD based MD method [6].

### 4.1 MD with BDD/ZDD

BDD and Zero-suppressed BDD (ZDD) are the compressed data structures for set families. In many cases, BDD and ZDD require smaller memory space for storing families and calculates values of combinatorial operations faster. The algorithm for MD using BDD/ZDD consists of 4 steps as follows:

**Step 1.** Construct a ZDD $p$ that corresponds to a set family $\phi$;
**Step 2.** Compute a BDD $q$ that corresponds to the HSs of $S(p)$, where $S(p)$ is the family of sets stored in $p$;

**Step 3.** Minimize $q$ into a ZDD $r$ that corresponds to the MHSs of $S(q)$;
**Step 4.** Output the $S(r)$ from $r$.

The algorithm utilizes fast techniques [6] for constructing and minimizing ZDDs (i.e., Step 1 and 3). In Step 2, it computes the intermediate BDD from an input ZDD in the recursive manner, as explained in the following example.

*Example 5.* Let the set family $\phi$ be $\{\{a,b\},\ \{a,c\},\ \{b,c,d\}\}$. The left diagram in Fig. 5 represents the ZDD of $\phi$. Every node $n$ is labeled by its value $v_n$, and has two children nodes $n.ph$ and $n.pl$ each of which is linked by the solid and dotted lines, respectively. The ZDD whose rooted note is $n.ph$ (*resp.* $n.pl$) corresponds to the family of sets with (*resp.* without) $n$. Note that $\top$ and $\bot$ mean true and false, respectively. In the practical point of view, every node $n$ in BDD/ZDD is managed in a hash table whose value is the tuple $\langle v_n, n.pl, n.ph\rangle$.

Now, we denote by $S(n)$ the family of sets stored in the ZDD whose root node is $n$. For instance, $S(a)$ is $\phi$ itself, $S(a.pl)$ is $\{\{b,c,d\}\}$, and $S(a.ph)$ is $\{\{b\},\{c\}\}$. Then, we notice that every hitting set $E$ of $\phi$ satisfies the condition that if $E$ contains $a$, then $E$ is a hitting set of $S(a.pl)$, otherwise it should be a hitting set of both $S(a.ph)$ and $S(a.pl)$. We denote by $HIT(a)$ the BDD corresponding to the hitting sets of $S(a)$ for the root node $a$. Then, $HIT(a)$ can be obtained in such a way that we recursively compute $HIT(a.ph)$ and $HIT(a.pl)$, merge them into the BDD $u$ by applying the "and" operation, and construct a new BDD whose root node corresponds to the tuple $\langle v_a, u, HIT(a.pl)\rangle$ (See Fig. 5).

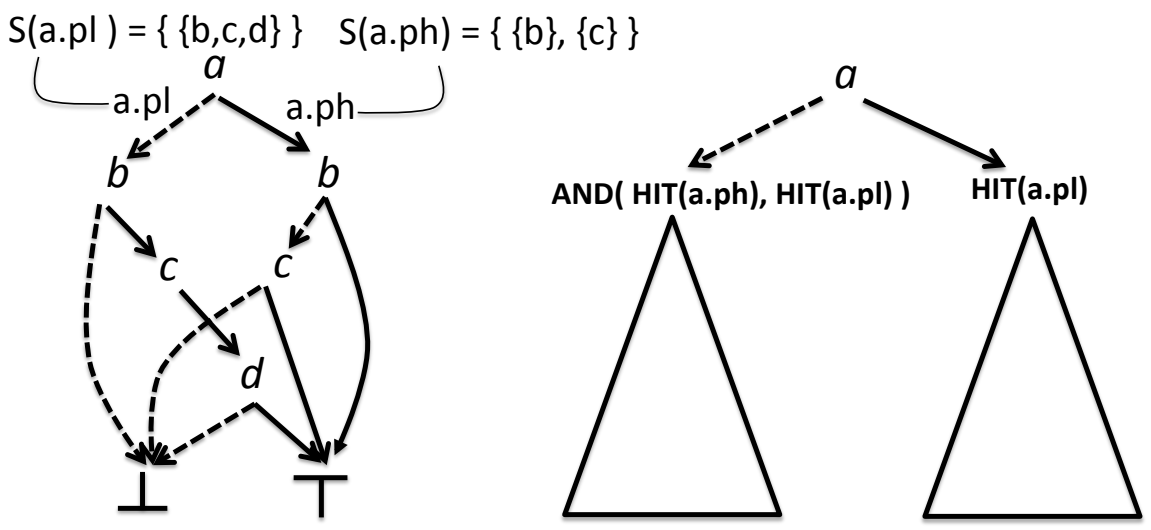

**Fig. 5.** MD computation with BDD/ZDD

This algorithm performs the MD computation only in the compressed data structures BDD and ZDD. For this feature, it can solve large-scale problems more efficiently than the MD method based on enumeration trees. For a preliminary experiment, we compared the execution time between those two methods using inductive learning problems that need dualization [17], as shown in Table 3.

The first line denotes the number of prime implicates for each problem. We put the label "T.O." if the execution time is over 60 sec. The BDD/ZDD based method solves all the problems faster than the other with enumeration trees.

### 4.2 A new algorithm for NMD with BDD/ZDD

This result motivates us to consider NMD using the BDD/ZDD based MD method. Given a CNF formula $\phi$, we can compute two ZDDs $p$ and $p_t$ that corresponds to $M(\phi)$ and $M(\phi \cup Taut(\phi))$, respectively. After removing tautologies from $p$ and $p_t$, we can use $p_t$ (corresponding to the bottom theory) to minimize $p$ by Theorem 3, as described in Fig. 6.

| Problems | animal | plus | arch. | multiple | oddeven9 | oddeven21 | oddeven31 |
|---|---|---|---|---|---|---|---|
| Num of PIs | 16 | 4.1E+3 | 3.3E+4 | 1.7E+7 | 128 | 1.1E+5 | 1.1E+9 |
| Enum. tree based one | 46 msec | 149 msec | 359 msec | T.O. | 37 msec | 627 msec | T.O. |
| BDD/ZDD based one | 17 msec | 18 msec | 18 msec | 18 msec | 17 msec | 18 msec | 18 msec |

**Table 3.** Comparison between the two MD methods

**Begin**
compute two ZDDs $p$ and $p_t$ from $\phi$;
/* $p$ and $p_t$ correspond to $M(\phi)$ and $M(\phi \cup Taut(\phi))$, respectively. */
remove the tautologies from $p$ and $p_t$;
while $p_t$ is not empty do
  select a set $e$ stored in $p$;
  $u_e := p_t.factor1(e)$; /* the sets stored in $p_t$ each of which contains $e$ */
  if $u_e$ is empty then remove $e$ from $p$;
  else $p_t := p_t$ - $u_e$;
return $p$;
**End**

**Fig. 6.** A new algorithm for NMD with BDD/ZDD

*Example 6.* Recall Example 3. Suppose that we have already obtained the ZDDs $p$ and $p_t$ and removed all the tautologies from them. We first select a set $e$ from $p$. Let $e$ be $\{\overline{x_1}, \overline{x_3}\}$. Then, we have $u_e = \{\{\overline{x_1}, x_2, \overline{x_3}\}, \{\overline{x_1}, \overline{x_2}, \overline{x_3}\}\}$. Since $u_e$ is not empty, we remove $u_e$ from $p_t$. In turn, we select the set $e = \{\overline{x_2}, x_3\}$. We have $u_e = \{\{x_1, \overline{x_2}, x_3\}, \{\overline{x_1}, \overline{x_2}, x_3\}\}$. Since it is not empty, we remove $u_e$ from $p_t$. We next select $e = \{x_1, x_2\}$. Then $u_e$ corresponds to $p_t$. By removing $u_e$ from $p_t$, $p_t$ becomes empty. Accordingly the other sets in $p$ are now redundant to be removed. Hence, the output $\{\{\overline{x_1}, \overline{x_3}\}, \{\overline{x_2}, x_3\}, \{x_1, x_2\}\}$ is obtained. Note that those processes are done in the ZDD (See Fig. 7).

Our proposal performs NMD with two steps: it first computes the ZDD corresponding to the prime implicates by the BDD/ZDD based MD method, and then minimizes it. Unlike the previous work [6], it requires the task of minimizing the ZDD $p$ with respect to another ZDD $p_t$.

There is a well-known method, called *Minato-Morreale* algorithm [15], for minimization of ZDDs. This algorithm applies so-called *recursive operator* in ZDD. The key technique is the hash-based cache to store the results of each recursive call. In contrast, our proposal uses the bottom theory for minimization of ZDDs. Once we construct the ZDD corresponding to the bottom theory, the minimization can be done only by applying the difference operator in ZDDs.

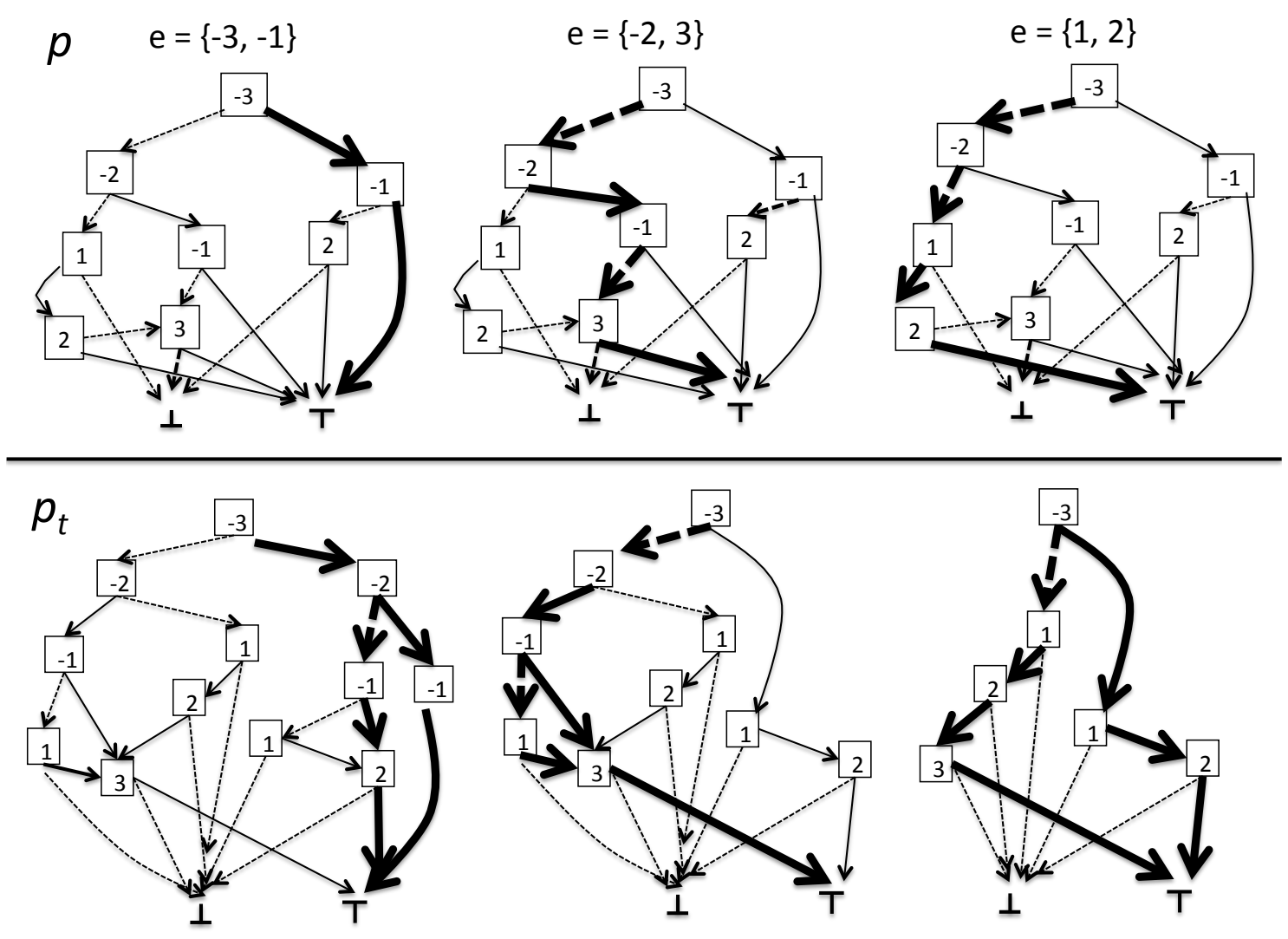

**Fig. 7.** NMD computation with BDD/ZDD

**Theorem 4.** *The algorithm can compute an output of NMD.*

*Proof.* The algorithm does not specify how to select the set $e$ from $p$. Suppose that an output $\psi$ is $\{e_1, e_2, \ldots, e_n\}$. By Theorem 3, $\psi$ minimally subsumes the bottom theory. Hence, each $e_i$ has its own clause in the bottom theory that is subsumed by $e_i$ but not subsumed by another. Accordingly, it is sufficient for deriving $\psi$ to select each $e_i$ $(1 \leq i \leq i)$ in order. □

## 5 Conclusion and future work

Recently, it is made known that any NMD can be logically reduced into two equivalent MD problems. Base on this reduction technique, we have investigated NMD based on practically fast MD computation. We focus on two MD methods with enumeration trees and BDD/ZDDs. First, we propose a pre-filtering algorithm CoCo1 that can eliminate redundant clauses through generating the prime implicates in MD computation. Using randomly generated instances, we have empirically shown that CoCo1 efficiently works in such problems that contain many redundant clauses. We next propose a new algorithm with the BDD/ZDD based MD method. Unlike the previous work, the key feature is to use the bottom theory for minimizing the ZDD corresponding to the prime implicates.

In terms of NMD with enumeration trees, it is important future work to perform furthermore experiments using larger-scale problems. In terms of NMD with BDD/ZDDs, it is necessary to formally analyze the condition to satisfy the minimality of outputs obtained by the proposed algorithm. We also intend to empirically compare our minimization technique with the previously proposed ones like Minato-Morreale algorithm.

## References

1. Eiter, T., Makino, K.: Abduction and the dualization problem. In: Proceedings of the 6th Int. Conf. on Discovery Science. Volume 2843 of LNCS. (2003) 1–20
2. Eiter, T., Makino, K., Gottlob, G.: Computational aspects of monotone dualization: A brief survey. Discrete Applied Mathematics **156** (2008) 2035–2049
3. Fredman, M., Khachiyan, L.: On the complexity of dualization of monotone disjunctive normal forms. Algorithms **21** (1996) 618–628
4. Uno, T.: A practical fast algorithm for enumerating minimal set coverings. In: Proceedings of the 83rd SIGAL Conf. of the Information Processing Society of Japan. (2002) 9–16
5. Murakami, K.: Efficient algorithms for dualizing large-scale hypergraphs. In: Proceedings of the 15th Meeting on Algorithm Engineering and Experiments, ALENEX 2013. (2013) 1–13
6. Toda, T.: Hypergraph Transversal Computation with Binary Decision Diagrams. In: Proceedings of the 12th Int. Sympo. on Experimental Algorithms (SEA2013). Volume 7933 of LNCS. (2013) 91–102
7. Avis, D., Fukuda, K.: Reverse search for enumeration. Discrete Applied Mathematics **1996** (1996) 21–46
8. Yamamoto, Y., Iwanuma, K., Inoue, K.: Non-monotone Dualization via Monotone Dualization. In: Proceedings of the 22nd Int. Conf. on Inductive Logic Programming. Volume 975 of CEUR. (2012) 74–79
9. Inoue, K.: Induction as consequence finding. Machine Learning **55(2)** (2004) 109–135
10. Yamamoto, A.: Hypothesis finding based on upward refinement of residue hypotheses. Theoretical Computer Science **298** (2003) 5–19
11. DeRaedt, L.: Logical settings for concept-learning. Artificial Intelligence **95** (1997) 187–201
12. Sakama, C., Inoue, K.: Brave induction: a logical framework for learning from incomplete information. Machine Learning **76** (2009) 3–35
13. Rymon, R.: An SE-tree based prime implicant generation algorithm. Annals of Mathematics and Artificial Intelligence (1994) 351–366
14. Brayton, R.K., Sangiovanni-Vincentelli, A.L., McMullen, C.T., Hachtel, G.D.: Logic Minimization Algorithm for VLSI Synthesis. Kluwer Academic Publishers (1984)
15. Minato, S.: Zero-suppressed BDDs and their applications. Int. Journal on Software Tools for Technology Transfer **3** (2001) 156–170
16. Satoh, K., Uno, T.: Enumerating maximal frequent sets using irredundant dualization. In: Proceedings of the 6th Int. Conf. on Discovery Science. Volume 2843 of LNCS. (2002) 256–268
17. Yamamoto, Y., Inoue, K., Iwanuma, K.: Heuristic Inverse Subsumption in Full-clausal Theories. In: Proceedings of the 22nd Int. Conf. on Inductive Logic Programming. Volume 7842 of LNCS. (2012) 241–256